\documentclass[10pt,twocolumn,letterpaper]{article}
\usepackage{cvpr} 

\definecolor{cvprblue}{rgb}{0.21,0.49,0.74}
\usepackage[pagebackref,breaklinks,colorlinks,allcolors=cvprblue]{hyperref}

\usepackage{booktabs}
\usepackage{multirow}
\usepackage{subcaption}
\usepackage{amsmath}
\usepackage[table]{xcolor}
\usepackage{graphicx}
\usepackage{pifont}
\usepackage{acronym}
\usepackage{marvosym}

\usepackage[capitalize]{cleveref}
\crefname{section}{Sec.}{Secs.}
\Crefname{section}{Section}{Sections}
\Crefname{table}{Table}{Tables}
\crefname{table}{Tab.}{Tabs.}
\crefname{algocf}{Alg.}{Algs.}
\Crefname{algocf}{Algorithm}{Algorithms}

\makeatletter
\DeclareRobustCommand\onedot{\futurelet\@let@token\@onedot}
\def\@onedot{\ifx\@let@token.\else.\null\fi\xspace}
\def\eg{\emph{e.g}\onedot} 

\def\ie{\emph{i.e}\onedot}

\def\etc{\emph{etc}\onedot} 
\def\vs{\emph{vs}\onedot}

\def\etal{\emph{et al}\onedot}
\makeatother

\acrodef{llm}[LLM]{Large Language Models}
\acrodef{vlm}[VLM]{Vision-Language Model}
\acrodef{sota}[SOTA]{state-of-the-art}
\acrodef{cem}[CEM]{Cross-Entropy Method}
\acrodef{gt}[GT]{Ground-Truth}
\acrodef{cd}[CD]{Chamfer Distance}

\newcommand{\rot}{\mathbf{R}}
\newcommand{\trans}{\mathbf{t}}
\newcommand{\scenetree}{\mathcal{T}}
\newcommand{\objlist}{\mathcal{O}}
\newcommand{\masklist}{\mathcal{M}}
\newcommand{\objmesh}{\mathbf{M}}
\newcommand{\sceneraw}{\mathbf{S}^\text{raw}}
\newcommand{\sceneinit}{\mathbf{S}^\text{cano}}
\newcommand{\scene}{\mathbf{S}}
\newcommand{\agentseg}{A^\text{seg}}
\newcommand{\agentv}{A^\text{ver}}

\newcommand{\group}{\mathbf{P}}
\newcommand{\groupinit}{\mathbf{P}^\text{cano}}
\newcommand{\rotinit}{\mathbf{R}^\text{cano}}
\newcommand{\transinit}{\mathbf{t}^\text{cano}}
\newcommand{\groupk}{\group^{(k)}}
\newcommand{\groupidx}{g}
\newcommand{\groupbest}{\group^\star}

\newcommand{\elite}{\mathcal{E}}

\newcommand{\lossstab}{E_\text{stab}}
\newcommand{\losslay}{E_\text{layout}}
\newcommand{\losscoll}{E_\text{pen}}
\newcommand{\lossvel}{E_\text{vel}}

\newcommand{\wstab}{\lambda_{\mathrm{stab}}}
\newcommand{\wstabposdr}{\lambda_{\mathrm{pos}}}
\newcommand{\wpen}{\lambda_{\mathrm{pen}}}
\newcommand{\wlayout}{\lambda_{\mathrm{layout}}}
\newcommand{\wvel}{\lambda_{\mathrm{vel}}}

\newcommand{\simstep}{L}
\newcommand{\placed}{(k)}
\newcommand{\settled}{(k),\simstep}

\newcommand{\rawy}{Y'}

\def\ourmethod{REST3D\xspace}

\def\ourrepo{\url{https://shirleymaxx.github.io/REST3D/}\xspace}
\newcommand{\ourpagetext}{\href{https://shirleymaxx.github.io/REST3D/}{project page}\xspace}
\def\ourmethod{REST3D\xspace}
\newcommand{\ourrepotext}{\href{https://github.com/ShirleyMaxx/REST3D}{GitHub repo}\xspace}

\usepackage[ruled,linesnumbered]{algorithm2e}

\SetAlFnt{\small}
\SetAlCapFnt{\small}
\SetAlCapNameFnt{\small}
\SetAlCapHSkip{0pt}

\begin{document}

\title{\ourmethod: Reconstructing Physically Stable 3D Scenes from a Single Image}

\author{
Xiaoxuan Ma \quad
Jiashun Wang \quad
Nicolás Ugrinovic \quad
Yehonathan Litman \quad
Kris Kitani \\[1.5ex]
Carnegie Mellon University \\
}

\twocolumn[{%
    \renewcommand\twocolumn[1][]{#1}%
    \setlength{\tabcolsep}{0.0mm} 
    \newcommand{\sz}{0.125}  
    \maketitle
    \vspace{-1.2em}
    \begin{center}
     \newcommand{\teaserwidth}{\textwidth}
        \includegraphics[width=\linewidth]{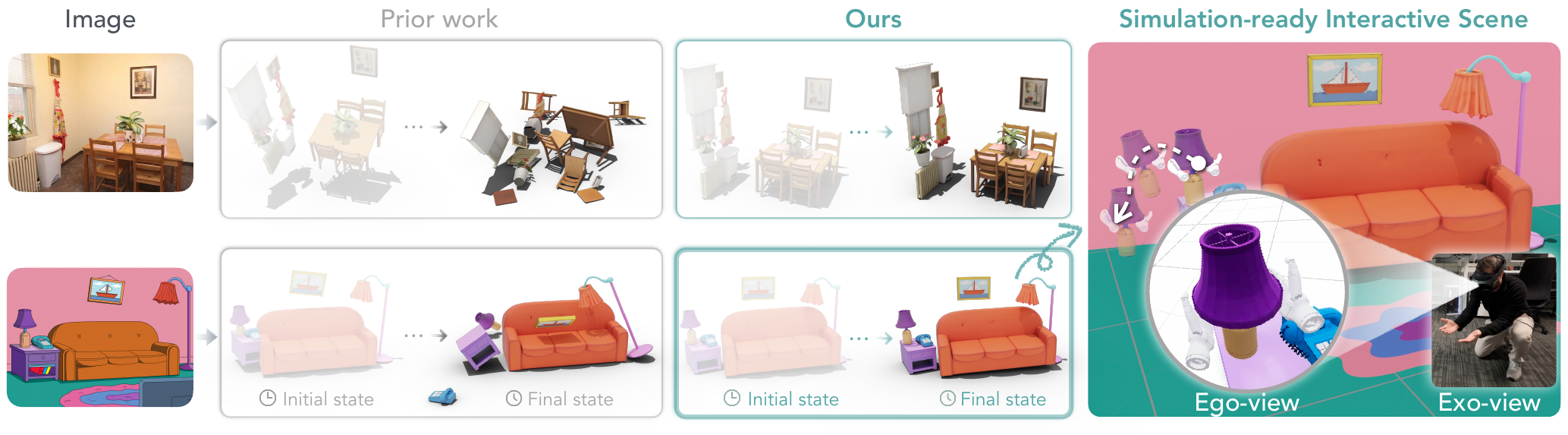}
    \vspace{-1.2em}
    \captionof{figure}{\textbf{We reconstruct physically stable 3D scenes from a single image, ensuring both visual consistency and physical plausibility.} Prior work \cite{chen2025sam3d} often produces physically implausible scenes, leading to unstable states when imported into a simulator \cite{makoviychuk2021isaac}, while ours obtains stable layouts that are ready for simulation, enabling seamless human–object interaction in VR environments. Project page: \ourrepo.} 
    \label{fig:teaser}
    \end{center}%
}]

\begin{abstract}
Reconstructing physically stable 3D scenes from a single RGB image enables casual images to be converted into simulation-ready digital assets for applications such as immersive interaction and content creation. However, existing single-image reconstruction methods fall short in capturing the physical structure of a scene. As a result, they often produce geometrically plausible but physically inconsistent results, including object floating and penetration, which lead to unstable behavior in physics simulations. Image-conditioned scene generation methods improve physical plausibility but often rely on strong scene priors, yielding plausible yet inaccurate object arrangements that fail to match the input image. We propose \ourmethod, a single-image reconstruction framework that can \underline{RE}construct physically \underline{ST}able \underline{3D} scenes by integrating physical scene understanding with physics-constrained refinement. We first introduce an agentic physical scene understanding technique that constructs a scene-tree representation capturing object physical states and inter-object relationships from a gravity-support perspective, providing a structural prior for reconstruction. Leveraging this structure, we initialize the scene using image-to-3D models, followed by scene-tree-guided alignment and physics-constrained optimization to resolve physical violations while preserving visual consistency with the input image. Experiments show that our method significantly reduces physical errors and improves simulation stability on both synthetic and real-world datasets while maintaining strong reconstruction quality. We further demonstrate the reconstructed scenes in VR-based human-object interaction, showing their potential for immersive applications.
\end{abstract}

\maketitle

\section{Introduction}
\label{sec:intro}

Reconstructing physically stable 3D scenes from a single RGB image enables the conversion of casual captures into simulation-ready digital assets for applications such as immersive interaction, content creation, gaming \cite{xia2025drawerdigitalreconstructionarticulation,wang2026contactguided}, \etc. Unlike multi-view or video inputs \cite{ni2024phyrecon}, the single-image setting is inherently more challenging due to limited visual observations. Nevertheless, a single image still contains rich information about the physical structure of the world such as object presence, geometry, appearance, and the physical contact between objects.

Despite rapid progress in single-image 3D reconstruction, existing methods \cite{factored3dTulsiani17,Ardelean2025Gen3DSR,meng2026scenegen,popov20eccv,Nie_2020_CVPR,chen2025sam3d,sautter20253dregen} often produce physically invalid results. Objects that should rest on supporting surfaces may float, while nearby objects may penetrate each other, leading to unstable behavior in physics simulation (see the collapsed final states in \cref{fig:teaser}). Retrieval-based approaches \cite{dai2024acdc} attempt to improve physical plausibility by assembling scenes using retrieved 3D assets, but are constrained by the coverage of the 3D asset database and often yield mismatched objects. More recently, conditional scene generation methods \cite{xia2026sage, wang2025tabletopgen, pfaff2026scenesmith} leverage agents to synthesize scenes from text or images. 
While capable of generating physically plausible layouts, these approaches often rely on strong learned scene priors, producing plausible but inaccurate object arrangements that fail to faithfully reproduce the observed scene.

In short, existing approaches either appear visually correct but fail to satisfy physical constraints, or remain physically valid but fail to recover the scene layout consistent with the input image. To address this, we propose \ourmethod, a single-image reconstruction framework that can \underline{RE}construct physically \underline{ST}able \underline{3D} scenes by integrating physical scene understanding with physics-constrained refinement (see \cref{fig:pipeline}). We first design an agentic physical scene understanding technique built on \ac{vlm}s to infer a scene-tree structure that captures object physical states and inter-object relationships (\eg, is vertically supported by X), enabling physically grounded reconstruction. Leveraging this structure, we initialize the scene using image-to-3D models \cite{chen2025sam3d}, and perform scene-tree-guided alignment and physics-constrained optimization to resolve physical errors while preserving visual consistency with the image. Specifically, we adopt a divide-and-conquer strategy to refine the scene layout and resolve physical errors by decomposing the scene into hierarchical groups according to the scene tree. We perform local refinement within each group, followed by global optimization over the entire scene to further refine the scene layout.

Experiments show that our method significantly reduces physical errors and outperforms \ac{sota} methods on both synthetic and real-world datasets in terms of reconstruction quality and physical stability, achieving an 83\%-point improvement in stability over \ac{sota} method \cite{chen2025sam3d}. Extensive ablation studies validate the effectiveness of the proposed scene-tree representation for physically stable reconstruction. Additionally, we implemented an interactive VR system that reconstructs an immersive and physically grounded 3D scene from a single image, enabling users to naturally interact with stable virtual objects through hand-based interactions. In summary, our contributions are as follows:
\begin{itemize}
    \item We introduce an agentic physical scene understanding technique that infers a scene-tree representation capturing object-level physical states and relationships, serving as the foundation for physically grounded reconstruction.

    \item Leveraging the scene-tree, we propose a framework that integrates scene understanding with physics-constrained optimization, correcting physical violations while preserving visual consistency with the input image.
    
    \item Our method achieves \ac{sota} physical stability and reconstruction quality on both synthetic and real-world data, enabling stable simulation and immersive interaction.
\end{itemize}

\section{Related Work}
\label{sec:related}

\subsection{3D Scene Reconstruction}
3D scene reconstruction has been studied from RGB-D scans or videos \cite{huang2026simulation,yu2026picasso,mcvay2025locate,avetisyan2020scenecad,huang2025literealitygraphicsready3dscene}, point clouds \cite{jiang2022ditto,Hsu2023DittoITH}, and multi-view RGB images or videos \cite{torne2024rialto, ni2024phyrecon, ni2025dprecon, xia2025drawerdigitalreconstructionarticulation, xia2025holoscene, xia2026simrecon,wang2026trisplat}. Despite their accuracy, they often require specialized sensing setups.

Single-image reconstruction has recently gained traction due to its practicality. Early methods \cite{factored3dTulsiani17, Nie_2020_CVPR, popov20eccv} use ResNet or 3D CNNs to predict voxel occupancy, but are limited to predefined object categories. Later approaches \cite{huang2025midi,Zhao_2025_ICCV_DepR,Ardelean2025Gen3DSR, meng2026scenegen, sautter20253dregen} leverage diffusion models or foundation models, \eg, TREELIS \cite{xiang2025structured}, to generate 3D objects and compose them into a complete scene. SAM3D \cite{chen2025sam3d} further improves image-to-3D reconstruction quality but mainly focuses on individual objects rather than holistic scenes. Recent works incorporate vision-language reasoning and retrieval into reconstruction \cite{chen2024urdformer,dai2024acdc,yin2026vision,jiayi2024singapo,le2024articulate}. DigitalCousins \cite{dai2024acdc} retrieves and composes similar 3D assets, but is constrained by the asset database and may produce mismatched reconstructions.

These single-image approaches mainly focus on visual reconstruction, while overlooking scene-level physical plausibility. 
Recent methods~\cite{huang2026simulation, yu2026picasso} address physical plausibility but require depth sensors and target small-scale scenes, while CAST~\cite{cast} and 3D-RE-GEN \cite{sautter20253dregen} consider only static physical relations without ensuring stability under full dynamics. In contrast, our method combines image-to-3D priors with explicit physical constraints for simulation-ready, room-level reconstruction from a single casual RGB image.

\subsection{3D Scene Generation}
With the rapid development of generative models and LLMs, recent works have shifted toward 3D scene synthesis. Early works \cite{wang2024architect,lin2024instructscene,yang2024holodeck} generate scenes from text descriptions. Recent methods incorporate multimodal reasoning and physical awareness, with some focusing on tabletop scenes \cite{lin2025pat3dphysicsaugmentedtextto3dscene,wang2025tabletopgen}. Pfaff \etal \cite{pfaff2025_steerable_scene_generation} project object translations to the nearest collision-free configurations while fixing orientations, limiting full 6-DoF correction and sometimes failing to converge. Further approaches \cite{yin2026vision,xia2026sage,pfaff2026scenesmith,chen2026scenefoundry,wang2026physcensis} explore agentic pipelines for scene generation from prompts. Among them, PhyScensis \cite{wang2026physcensis} conveys physics feedback through language, resulting in 3D spatial information loss. Other methods \cite{xia2026sage,wang2025tabletopgen} use intermediate images for generation, partially resembling reconstruction, but may rely on heuristics and offer limited control, potentially causing misalignment with images. In contrast, we use agent assistance for well-defined sub-problems rather than fully agent-driven generation, enabling more controllable, image-conditioned reconstruction.

\section{Method}
\label{sec:method}

\begin{figure*}[t]
  \centering
  \includegraphics[width=\linewidth]{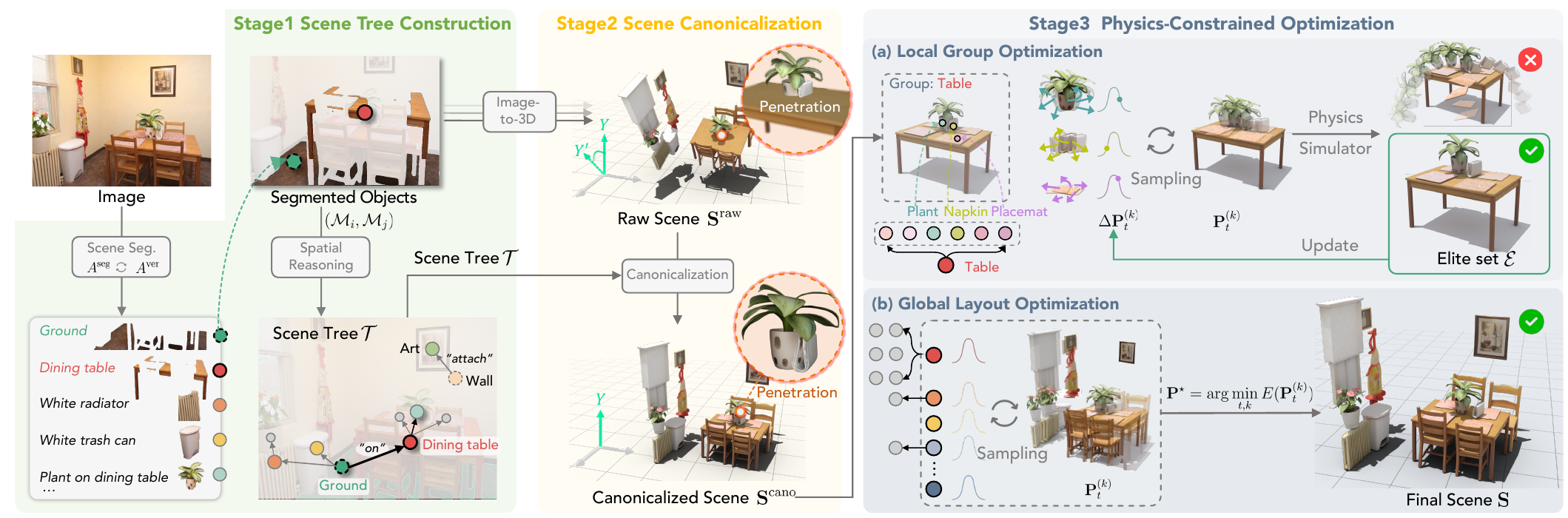}
  \caption{\textbf{\ourmethod overview.} Given a single RGB image $I$, \ourmethod reconstructs a physically stable 3D scene $\scene$ ready for physics simulation. It consists of three stages: \textbf{(1) Scene-Tree Construction} (\cref{sec:scenetree}) infers a scene tree $\scenetree$ encoding object spatial relationships; \textbf{(2) Scene Initialization and Canonicalization} (\cref{sec:init}) initializes a raw scene and performs scene-tree-guided canonicalization to produce a structured initial scene $\sceneinit$; and \textbf{(3) Physics-Constrained Optimization} (\cref{sec:opt}) refines object poses through simulation-based optimization under physical constraints, resulting in a physically stable scene $\scene$.
  }
  \label{fig:pipeline}
  \vspace{-0.8em}
\end{figure*}

Given a single RGB image $I$, our goal is to reconstruct a physically plausible 3D scene $\scene$, including object meshes and their corresponding world-frame layout ready for physics simulation.
\cref{fig:pipeline} illustrates our overall framework, which consists of three stages. First, \textbf{Scene-Tree Construction} (\cref{sec:scenetree}) infers a hierarchical scene tree that encodes object physical states and their spatial relationships. Next, \textbf{Scene Initialization and Canonicalization} (\cref{sec:init}) initializes a 3D scene and performs scene-tree-guided canonicalization to produce a structured initial layout. Finally, \textbf{Physics-Constrained Optimization} (\cref{sec:opt}) refines object poses through simulation-based optimization under physical constraints, resulting in a physically consistent scene while preserving the visual consistency.

\subsection{Scene Tree Construction}
\label{sec:scenetree}
To enable automatic reconstruction from arbitrary images, we decompose the process into three steps with specialized physical scene understanding agents: (i) open-vocabulary object list analysis for scene object identification; (ii) agentic instance segmentation for per-object masks; and (iii) scene-tree construction via spatial relationship reasoning (see Stage 1 in \cref{fig:pipeline}).

\vspace{0.8em}
\noindent\textbf{Open-vocabulary Object List Analysis.}
We first identify all objects in the input image. Specifically, we design a scene analyzer that leverages a \ac{vlm} (\ie, Gemini) to generate a structured open-vocabulary object list. Instead of producing coarse category labels like ``plant'', we encourage the model to distinguish individual instances within the same category by incorporating disambiguating attributes such as relative position (\eg, ``plant on dining table''). This results in an object list with descriptive attributes, denoted as $\objlist = \{\objlist_i\}$. See Supp. \cref{supp:sec:method} for more details.

\vspace{0.8em}
\noindent\textbf{Agentic Instance Segmentation.}
Given each object description $\objlist_i$, we could leverage SAM3~\cite{carion2026sam} to obtain its segmentation mask. However, due to occlusions and complexity in real-world scenes, directly using $\objlist_i$ as input to SAM3 may lead to inaccurate results. Therefore, we build an agentic system that integrates instance segmentation with visual-language reasoning, consisting of two specialized agents: a \textit{segmentation agent} $\agentseg$, which performs object segmentation, and a \textit{verifier agent} $\agentv$, which evaluates the quality of the segmented mask. At each iteration, $\agentseg$ refines $\objlist_i$ into a more detailed prompt tailored for $\agentseg$ to generate a candidate mask. The mask is then overlaid on the input image and passed to $\agentv$, which determines whether it matches the target object description. Based on the verifier’s feedback, $\agentseg$ further refines the prompt and repeats this process. The loop terminates when the verifier $\agentv$ confirms the mask or determines that the object is absent from the scene.
This iterative process produces a set of segmentation masks $\masklist = \{\masklist_i\}$ for all objects, where each mask $\masklist_i$ is associated with a unique object description $\objlist_i$. See Supp. \cref{supp:sec:method} for more details. 

\vspace{0.8em}
\noindent\textbf{Scene Tree Construction via Spatial Reasoning.}
Given object identifiers $\objlist$ and masks $\masklist$, we construct a support-relation scene tree $\scenetree$ using a spatial reasoning module. The tree is rooted at four canonical support nodes: \textit{ground}, \textit{wall}, \textit{ceiling}, and \textit{ground-wall}. A \ac{vlm} analyzes mask-overlaid object pairs $(\masklist_i,\masklist_j)$ from a gravity-aware perspective to infer the support parent and relation type (\ie, \textit{on}, \textit{hanging}, or \textit{attached to}). See an example scene tree $\scenetree$ in \cref{fig:pipeline} Stage 1. Objects supported by multiple surfaces, such as a radiator contacting both the ground and a wall, are assigned the composite \textit{ground-wall} parent. See Supp. \cref{supp:sec:method} for more details.

$\scenetree$ encodes each object’s support parent and hierarchical spatial relationships, providing a structured prior for layout initialization (\cref{sec:init}) and physics-constrained optimization (\cref{sec:opt}).

\subsection{Scene Initialization and Canonicalization}
\label{sec:init}
Given the scene tree $\scenetree$ and instance masks $\masklist$, we obtain a raw 3D reconstruction $\sceneraw$ and apply scene-tree-guided canonicalization for global orientation and coarse support constraints, yielding a physically meaningful initial scene $\sceneinit$ (Stage~2 in \cref{fig:pipeline}).

\vspace{0.8em}
\noindent\textbf{Object Reconstruction and Scene Initialization.}
We leverage an image-to-3D model, \ie SAM3D~\cite{chen2025sam3d}, to reconstruct each object meshes $\{\objmesh_i\}$ from their masks $\masklist$. We then use the estimated rotation, translation, and scale for each object from SAM3D to initialize scene $\sceneraw$ in the world frame. Despite providing an initial reconstruction, $\sceneraw$ is often physically inconsistent. The estimated global pose may be inaccurate and misaligned, with frequent inter-object collisions. For example, the table may appear floating, while the plant penetrates the tabletop in \cref{fig:pipeline} ($\sceneraw$). When directly imported into a physics simulator, such inconsistencies result in unstable dynamics. Note that our method is model-agnostic and can incorporate improved \ac{vlm} or image-to-3D methods.

\vspace{0.8em}
\noindent\textbf{Scene Canonicalization.}
To address these issues, we perform scene-tree-guided canonicalization to align the scene with gravity and enforce coarse support constraints. We first estimate the dominant vertical direction $\rawy$ from ground-supported objects and large furniture anchors, and rotate the scene to align $\rawy$ with the world $Y$-axis (opposite to gravity); see \cref{fig:pipeline}. We then traverse $\scenetree$ to adjust child positions along the vertical direction according to their support relations, resolving coarse vertical penetrations. This yields a more physically plausible initialization $\sceneinit$, while residual inconsistencies (\eg inter-object intersections) remain. We therefore introduce scene-tree-guided physics-constrained optimization (\cref{sec:opt}) to further resolve these physical errors.

\subsection{Physics-Constrained Optimization}
\label{sec:opt}
Starting from $\sceneinit$, we optimize object poses for physical stability under full-dynamics simulation while preserving its layout, using simulator feedback \cite{makoviychuk2021isaac} and a scene-tree-guided divide-and-conquer strategy. We first optimize local groups, then jointly refine their global layout (Stage~3 in \cref{fig:pipeline}).

\vspace{0.8em}
\subsubsection{Local Group Optimization}
With the scene tree $\scenetree$, we construct local groups via post-order traversal. For each non-root node $\groupidx \in \scenetree$ with at least one child, we define a group consisting of $\groupidx$ and its direct children, and optimize their poses using a physics-informed objective to obtain a physically plausible layout relative to the group root. Post-order traversal ensures that when a node is included in its parent’s group, its own group has already been optimized. Leaf nodes are excluded from local groups and optimized during the global layout optimization.

\vspace{0.8em}
\noindent\textbf{Formulation.}
We aim to estimate a physically plausible and spatially consistent layout for a local group $\groupidx$ in the scene $\sceneinit$. Each group consists of $N_\groupidx$ child objects, where each object is represented by a 6-DoF pose. We denote the initial group layout as $\groupinit = \{(\rotinit_i, \transinit_i)\}_{i=1}^{N_\groupidx}$, where $\rotinit$ and $\transinit$ are rotation and translation, respectively. Starting from $\groupinit$, we seek a set of small pose adjustments that lead to a physically stable state while preserving the original layout. We formulate this problem as a distribution-based optimization over such adjustments, where the objective is to minimize a physics-informed energy function $E$ that measures physical feasibility and layout consistency.

We adopt the \ac{cem} \cite{RUBINSTEIN199789,de2005tutorial}, a population-based stochastic optimization technique. \ac{cem} maintains a parametric Gaussian distribution over pose adjustments and iteratively refines it toward high-quality solutions. Specifically, at iteration $t$, we model the adjustment distribution applied to the canonical group layout $\groupinit$ as a Gaussian with mean $\boldsymbol{\mu}_t \in \mathbb{R}^{N_\groupidx \times 6}$ and diagonal covariance $\boldsymbol{\Sigma}_t$. We then draw $K$ samples:
\vspace{-0.2em}
\begin{align}
\small
\Delta \group_t^{(k)} \sim \mathcal{N}(\boldsymbol{\mu}_t, \boldsymbol{\Sigma}_t), \quad k=1,\ldots,K,
\vspace{-0.5em}
\end{align}
and obtain candidate layouts $\groupk_t$ by applying $\Delta \group_t^{(k)}$ to $\groupinit$.

We import all candidates into a physics simulator (Isaac Gym \cite{makoviychuk2021isaac}) and perform simulation to obtain the resulting states. Based on the simulated outcomes, we compute a physics-based energy $E(\groupk_t)$ for each candidate. We select an elite set $\elite_t$ consisting of the top $\lceil \rho K \rceil$ lowest-energy samples, where $\rho \in [0,1]$ is a ratio parameter. Then the distribution is updated via moment matching over the corresponding adjustment samples:
\[
\small
\boldsymbol{\mu}_{t+1} \leftarrow \frac{1}{|\elite_t|} \sum_{k \in \elite_t} \Delta \group_t^{(k)}, \quad
\boldsymbol{\Sigma}_{t+1} \leftarrow \frac{1}{|\elite_t|} \sum_{k \in \elite_t} (\Delta \group_t^{(k)} - \boldsymbol{\mu}_{t+1})^2.
\]
This process is repeated for $T$ iterations, steering the sampling distribution toward physically stable and geometrically consistent layouts. Finally, we select the best one $\groupbest = \arg\min_{t, k} E(\groupk_t)$ as the final layout for the local group. See \cref{fig:pipeline} Stage 3 (a).

\vspace{0.8em}
\noindent\textbf{Energy Function.}
We leverage a physics engine \cite{makoviychuk2021isaac} to evaluate the energy $E$ of each candidate $\groupk$ by placing objects into the simulator using their sampled poses and running forward simulation. Based on the resulting simulation states, we define the energy function $E(\groupk)$ that consists of multiple terms:
\begin{equation}
\small
  E = \wstab \lossstab + \wvel \lossvel + \wpen \losscoll + \wlayout \losslay,
  \label{eq:energy_decomp}
\end{equation}
where $\lossstab$ and $\lossvel$ measure post-simulation stability, $\losscoll$ penalizes geometric penetration, and $\losslay$ measures consistency with scene $\sceneinit$. $\wstab$, $\wvel$, $\wpen$, and $\wlayout$ are hyperparameters.

\textit{Stability energy} $\lossstab$ measures how much objects drift from their initial placements after forward simulation. Specifically, we place each object $i$ according to its candidate pose $(\rot_i^{\placed}, \trans_i^{\placed})$ and run the simulator for $\simstep$ steps to obtain the post-simulation poses $(\rot_i^{\settled}, \trans_i^{\settled})$ for each object. We then measure the deviation between the initial and final states:
\begin{equation}
\small
  \lossstab = \sum_i \bigl(
    \|\trans_i^{\settled} - \trans_i^{\placed}\|
    + d(\rot_i^{\settled}, \rot_i^{\placed})\bigr),
  \label{eq:Estab}
\end{equation}
\vspace{-0.3em}
where $d(\cdot,\cdot)$ denotes the geodesic distance between rotations, implemented via quaternion distance. We further incorporate the measured \textit{velocity energy} $\lossvel$ at an intermediate simulation step $\tau$ as an early indicator of instability:
\begin{equation}
\small
  \lossvel = \sum_i \|\mathbf{v}_i\|,
  \label{eq:Evel}
\end{equation}
where $\mathbf{v}_i$ denotes the linear velocity of object $i$ at step $\tau$. A high velocity suggests that the object is still undergoing dynamic interactions (\eg, collisions) and has not yet settled to a stable state.

\textit{Penetration energy} $\losscoll$ penalizes geometric overlap at both placement and settlement:
\begin{equation}
\small
  \losscoll = \losscoll^{\placed} + \losscoll^{\settled},
  \label{eq:Epen}
\end{equation}
where $\losscoll^{\placed}$ counts pairwise convex-hull intersections at placement using the Gilbert–Johnson–Keerthi distance algorithm \cite{gilbert1988fast}, and $\losscoll^{\settled}$ counts intersections after $\simstep$-step settlement.
\textit{Layout energy} $\losslay$ enforces consistency of the settled scene with $\sceneinit$:
\begin{equation}
\small
  \losslay = \sum_i \bigl(
    \wstabposdr \, \|\trans_i^{\settled} - \transinit_i\|
    + d\big(\rot_i^{\settled},\, \rotinit_i)\big)
  \bigr).
  \label{eq:Elayout}
\end{equation}
This term prevents degenerate solutions that disperse objects excessively for physical stability, deviating too much from original layout. \\

\subsubsection{Global Group Optimization.}
After independently optimizing all local groups, we treat each optimized group as a rigid unit and jointly optimize it with objects whose parents are \textit{ground} or \textit{ground-wall}, forming a higher-level scene graph. We apply the same \ac{cem}-based physics-informed formulation with $T$ iterations and energy $E$. During simulation, each group is expanded into its constituent objects, whose poses are obtained by composing the optimized group-root poses with the local poses $\groupbest$. Finally, we heuristically place objects attached to wall or ceiling nodes to ensure collision-free placement while preserving layout consistency, yielding the final scene $\scene$. See Supp. \cref{supp:sec:placewall} for pseudocode and details.

\begin{table*}[t]
\centering
\caption{\textbf{Comparison with SOTA methods.} ``---'' denotes not applicable due to no GT mesh. SAM3D $\dagger$ indicates scene-level reconstruction (see \cref{sec:expset}).}
\vspace{-0.8em}
\label{tab:expsota}
\resizebox{\linewidth}{!}{
\setlength{\tabcolsep}{2mm}
\begin{tabular}{ll c cccccc ccc}
\toprule
\multirow{2}{*}{} & \multirow{2}{*}{Method} & Failure
& \multicolumn{5}{c}{Physical Metrics} & \multicolumn{3}{c}{Geometric Metrics} \\
\cmidrule(lr){4-8} \cmidrule(lr){9-11}
& & Rate (\%)$\downarrow$
& Coll. Rate (\%)$\downarrow$ 
& Stable Rate (\%)$\uparrow$
& Pos. Drift (m)$\downarrow$
& Lin. Vel. (m/s)$\downarrow$
& Ang. Vel. (rad/s)$\downarrow$
& CD$\downarrow$ 
& F-Score$\uparrow$ 
& B-IoU$\uparrow$ \\
\midrule
\multirow{8}{*}{\rotatebox{90}{Replica}}
& DigitalCousins \cite{dai2024acdc}       & \underline{10.0} & 25.0 & \underline{58.3} & \underline{0.217} & \underline{0.886} & \underline{1.084} & 0.0578 & 0.615 & 0.12 \\
& Gen3DSR \cite{Ardelean2025Gen3DSR}      & \textbf{0.0} & 61.2 & 7.5 & 0.836 & 3.017 & 1.699 & 0.0726 & 0.606 & 0.10 \\
& MIDI \cite{huang2025midi}               & \underline{10.0} & 48.8 & 25.9 & 0.441 & 1.715 & 2.253 & 0.1544 & 0.502 & 0.10 \\
& SceneGen \cite{meng2026scenegen}        & \textbf{0.0} & 83.4 & 15.6 & 1.182 & 3.329 & 4.977 & 0.0532 & 0.662 & 0.15 \\
& DepR \cite{Zhao_2025_ICCV_DepR}         & \textbf{0.0} & 41.0 & 2.9 & 0.637 & 1.672 & 1.739 & 0.0918 & 0.601 & 0.19 \\
& SAM3D $\dagger$ \cite{chen2025sam3d}    & \textbf{0.0} & 48.5 & 2.5 & 1.013 & 2.019 & 8.168 & \underline{0.0139} & \textbf{0.891} & \textbf{0.39} \\
& 3D-RE-GEN \cite{sautter20253dregen}     & 40.0 & \underline{23.7} & 14.1 & 0.861 & 3.167 & 19.025 & 0.0643 & 0.582 & 0.13 \\
& \textbf{Ours}                           & \textbf{0.0} & \textbf{6.7} & \textbf{96.2} & \textbf{0.037} & \textbf{0.169} & \textbf{0.539} & \textbf{0.0109} & \underline{0.868} & \underline{0.38} \\

\midrule

\multirow{8}{*}{\rotatebox{90}{ScanNet++}}
& DigitalCousins \cite{dai2024acdc}       & \underline{33.3} & \underline{20.4} & \underline{45.4} & \underline{0.481} & \underline{1.825} & \underline{2.452} & 0.087 & 0.539  & 0.07 \\
& Gen3DSR \cite{Ardelean2025Gen3DSR}      & 66.7 & 77.9 & 1.7 & 3.654 & 7.966 & 12.339 & 0.078 & 0.546  & 0.04\\
& MIDI \cite{huang2025midi}            & 0.0  & 47.5 & 27.2 & 0.712 & 2.266 & 3.317  & 0.099 & 0.524 & 0.08 \\
& SceneGen \cite{meng2026scenegen}        & 50.0 & 80.9 & 5.3 & 10.035 & 29.771 & 27.762 & 0.052 & 0.635  & 0.12 \\
& DepR \cite{Zhao_2025_ICCV_DepR}      & 25.0 & 46.8 & 4.4  & 0.575 & 2.365 & 2.057  & 0.094 & 0.498 & 0.12 \\
& SAM3D $\dagger$ \cite{chen2025sam3d}            & \textbf{0.0} & 39.1 & 4.6 & 0.948 & 2.096 & 8.121 &  \textbf{0.015} & \textbf{0.816}  & \underline{0.19} \\
& 3D-RE-GEN \cite{sautter20253dregen}  & 50.0 & 27.1 & 17.2 & 0.396 & 1.736 & 17.249 & 0.121 & 0.496 & 0.10 \\
& \textbf{Ours}                          & \textbf{0.0} & \textbf{5.9} & \textbf{93.6} & \textbf{0.080} & \textbf{0.159} & \textbf{1.039} & \underline{0.019} &  \underline{0.807} & \textbf{0.20} \\

\midrule

\multirow{8}{*}{\rotatebox{90}{Custom}}
& DigitalCousins \cite{dai2024acdc}       & 48.0 & \underline{25.0}                        & \underline{50.2} & \underline{0.911} & 2.196 & \underline{2.158} & --- &--- & --- \\ 
& Gen3DSR \cite{Ardelean2025Gen3DSR}      & 40.0 & 68.2                         & 7.2 & 2.756 & 6.493 & 9.009 & --- &--- & --- \\    
& MIDI \cite{huang2025midi}            & 0.0  & 46.6 & 22.0 & 0.404 & 2.143 & 2.998  & --- & --- & --- \\
& SceneGen \cite{meng2026scenegen}        & \underline{16.0} & 78.4             & 2.9 & 15.626 & 27.475 & 29.499 & --- &--- & --- \\   
& DepR \cite{Zhao_2025_ICCV_DepR}      & 8.0  & 26.5 & 17.2 & 0.355 & 1.604 & 3.694  & --- & --- & --- \\
& SAM3D $\dagger$ \cite{chen2025sam3d}            & \textbf{0.0} &        45.4       & 16.0 & 1.189 & \underline{1.929} & 11.801 & --- &--- & --- \\     
& 3D-RE-GEN \cite{sautter20253dregen}  & 36.0 & 28.5 & 28.9 & 0.412 & 1.891 & 52.006 & --- & --- & --- \\
& \textbf{Ours}                          & \textbf{0.0} & \textbf{1.2} & \textbf{95.5} & \textbf{0.017} & \textbf{0.140} & \textbf{0.468} & --- &--- & --- \\

\bottomrule
\end{tabular}}
\vspace{-0.5em}
\end{table*}
\section{Experiments}
\label{sec:exp}

\subsection{Setup}
\label{sec:expset}
\noindent\textbf{Baselines.}
We compare against 7 \ac{sota} methods \cite{dai2024acdc,Ardelean2025Gen3DSR,huang2025midi,meng2026scenegen,Zhao_2025_ICCV_DepR,chen2025sam3d,sautter20253dregen} that take a single image as input for scene reconstruction. All methods are evaluated using their publicly available code and pretrained models. Note that SAM3D is designed for single-object reconstruction. Here, we extend it by integrating our Stage 1 to enable scene-level reconstruction, \ie, $\sceneraw$ (see \cref{sec:init}), marked with $\dagger$ in the following tables and figures.

\begin{figure*}[t]
  \centering
  \includegraphics[width=.9\linewidth]{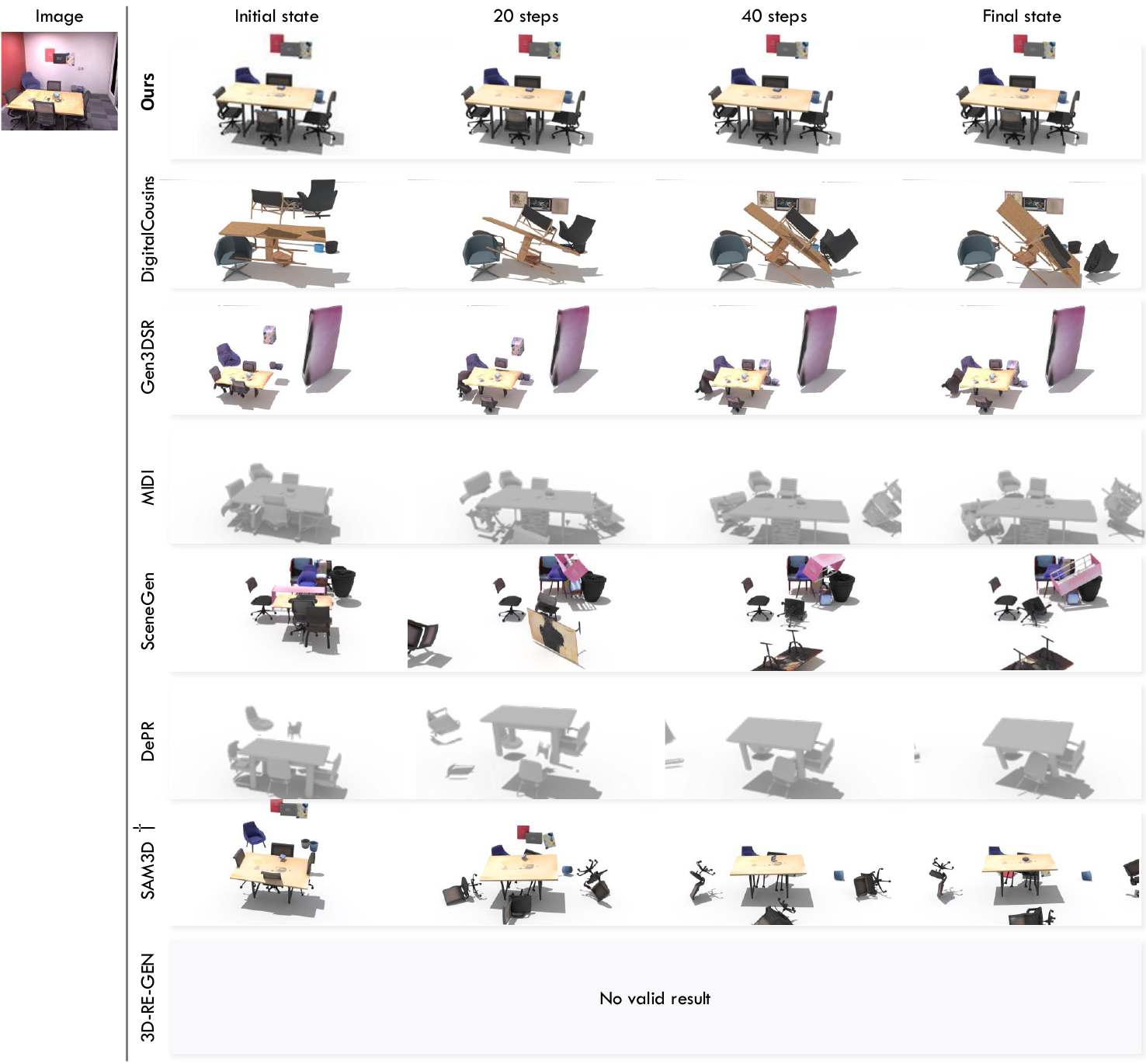}
  \caption{\textbf{Comparison of simulation processes for reconstructed scenes.} We visualize the simulation process of scenes reconstructed by different methods in a physics simulator (Isaac Gym). MIDI and DepR do not reconstruct textures, while 3D-RE-GEN fails to produce a reconstruction in this case due to unsuccessful point cloud extraction.} Image source: Replica \cite{replica19arxiv}.
  \label{fig:exp_sota}
\end{figure*}

\begin{figure}[t]
  \centering
  \includegraphics[width=\linewidth]{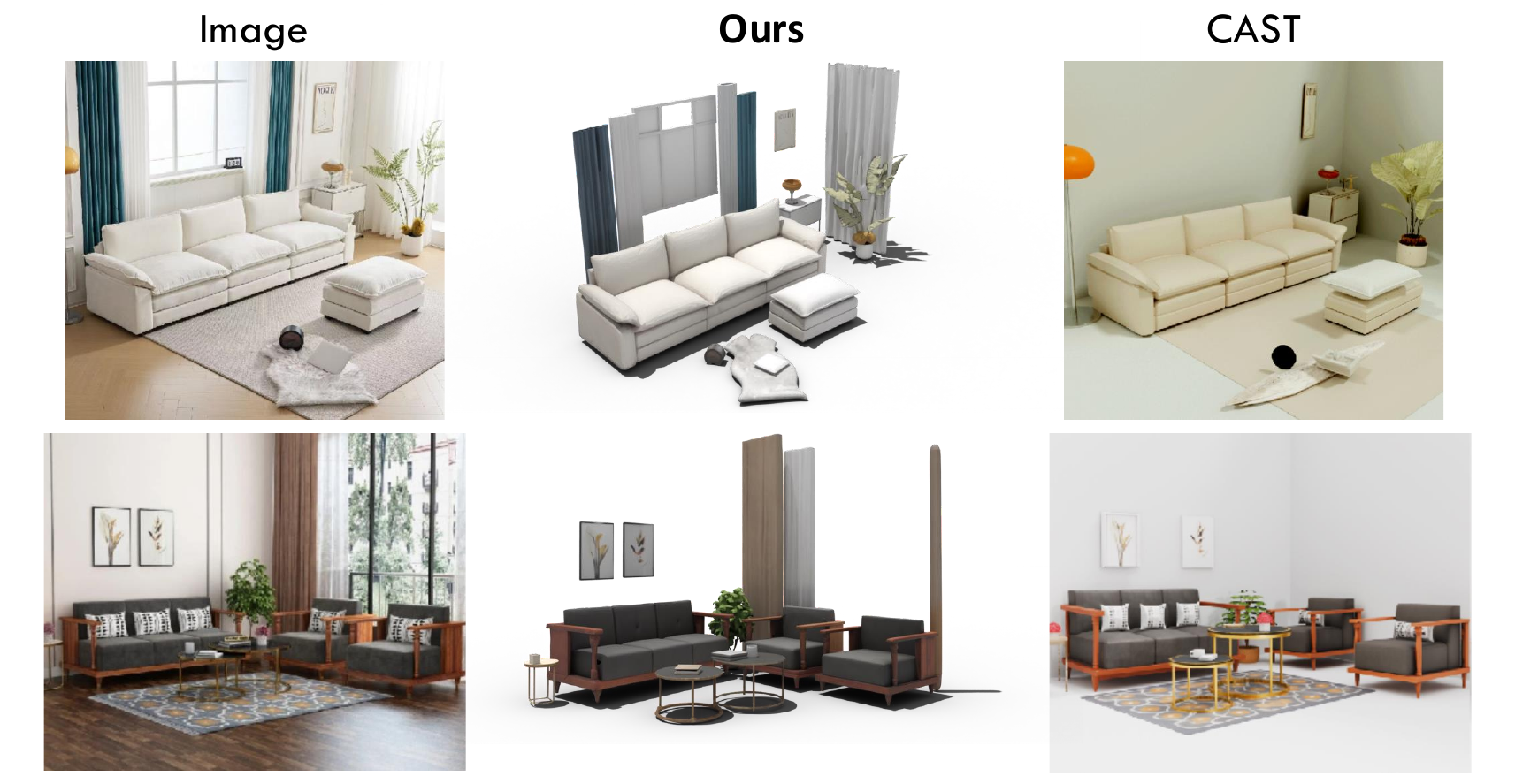}
  \caption{\textbf{Visual comparison with CAST~\cite{cast}.}}
  \label{fig:exp_cast}
\end{figure}

\vspace{0.8em}
\noindent\textbf{Datasets.}
We evaluate on both synthetic Replica dataset \cite{replica19arxiv} (10 scenes) and real-world ScanNet++ dataset \cite{yeshwanth2023scannetpp} (12 scenes), covering diverse scenes such as meeting rooms, classrooms, and offices. In addition, we construct a custom set of 25 more challenging scenes (denoted as \textit{Custom}), including bedrooms, living rooms, and even cartoon-style scenes, to evaluate robustness for a casual single image.

\vspace{0.8em}
\noindent\textbf{Metrics.}
We report the \textit{failure rate}, \ie, the proportion of methods that fail to reconstruct any valid scene. We further evaluate the physical plausibility of the reconstructed scenes by measuring the \textit{collision rate}, \textit{stability rate}, \textit{position drift} between the initial placement and the final settled state, as well as the \textit{peak linear and angular velocities}. For the Replica and ScanNet++ datasets, which provide \ac{gt} 3D scene meshes, we additionally evaluate geometric reconstruction accuracy using \ac{cd}, F-score@0.05 (F-Score), and B-IoU, following \cite{dai2024acdc,chen2025sam3d,meng2026scenegen}. Before evaluation, we apply ICP alignment \cite{besl1992method} to align the reconstructed scene with \ac{gt}. Please refer to Supp. \cref{supp:sec:metric} for detailed metric definitions.

\subsection{Implementation Details}
All \ac{vlm}s used in Stage 1 are Gemini 3 Flash. The instance segmentation agent calls SAM3 \cite{carion2026sam} as a tool. We use SAM3D \cite{chen2025sam3d} for image-to-3D. For Stage 3, we use Isaac Gym \cite{makoviychuk2021isaac} as the simulator and follow real-world physics as much as possible, \eg, setting gravity to $-9.8\,\mathrm{m/s^2}$. For \ac{cem} optimization, we run two episodes for both local and global optimization, each consisting of $T=15$ iterations. At each iteration, we sample $K=2048$ candidates and select the top $\rho=0.025$ as elite candidates. We then perform $L=60$-step simulation. For the coefficients, we set $\wstab = \wvel = \wlayout = 1$, $\wpen = 0.5$, and $\wstabposdr = 6$. 
Our method takes 25.8 min/scene, including 10.4 min for physics optimization. See Supp. \cref{supp:sec:expdetail} for more details and \cref{supp:sec:expresults} for runtime breakdown, API costs, and comparisons with baselines.

\subsection{Comparison with SOTAs}
\label{sec:exp_sota}
\noindent\textbf{Quantitative Results.}
Table~\ref{tab:expsota} reports quantitative results of our method and baselines. Our method achieves the best physical plausibility and reconstruction quality. On real-world ScanNet++ scenes and our custom casual images, retrieval-based methods such as DigitalCousins~\cite{dai2024acdc} fail to produce valid reconstructions in nearly half of the cases. In contrast, both SAM3D $\dagger$ and our method successfully handle all cases, benefiting from our agentic scene analysis pipeline and strong image-to-3D priors. Unlike MIDI, DepR, and 3D-RE-GEN, which use single-pass detection with fixed thresholds and no verification, our agentic pipeline iteratively verifies detections, reducing failure rate; see Supp. \cref{supp:sec:expresults} for detection accuracy comparison. Nevertheless, all baselines exhibit poor physical plausibility due to severe interpenetration and instability, with significant post-simulation position drift, leading to high collision rates and large linear and angular velocities.

For geometric metrics, our method and SAM3D are comparable. However, SAM3D may contain geometric errors, while our optimization corrects such errors and separates inter-penetrating objects to enforce physical plausibility, introducing small displacements from the GT and thus slightly worse F-Score. This reveals that geometric metrics do not account for scene-level physical validity. We provide a typical case analysis in Supp. \cref{supp:sec:expresults}.

\begin{figure*}[t]
  \centering
  \includegraphics[width=.98\linewidth]{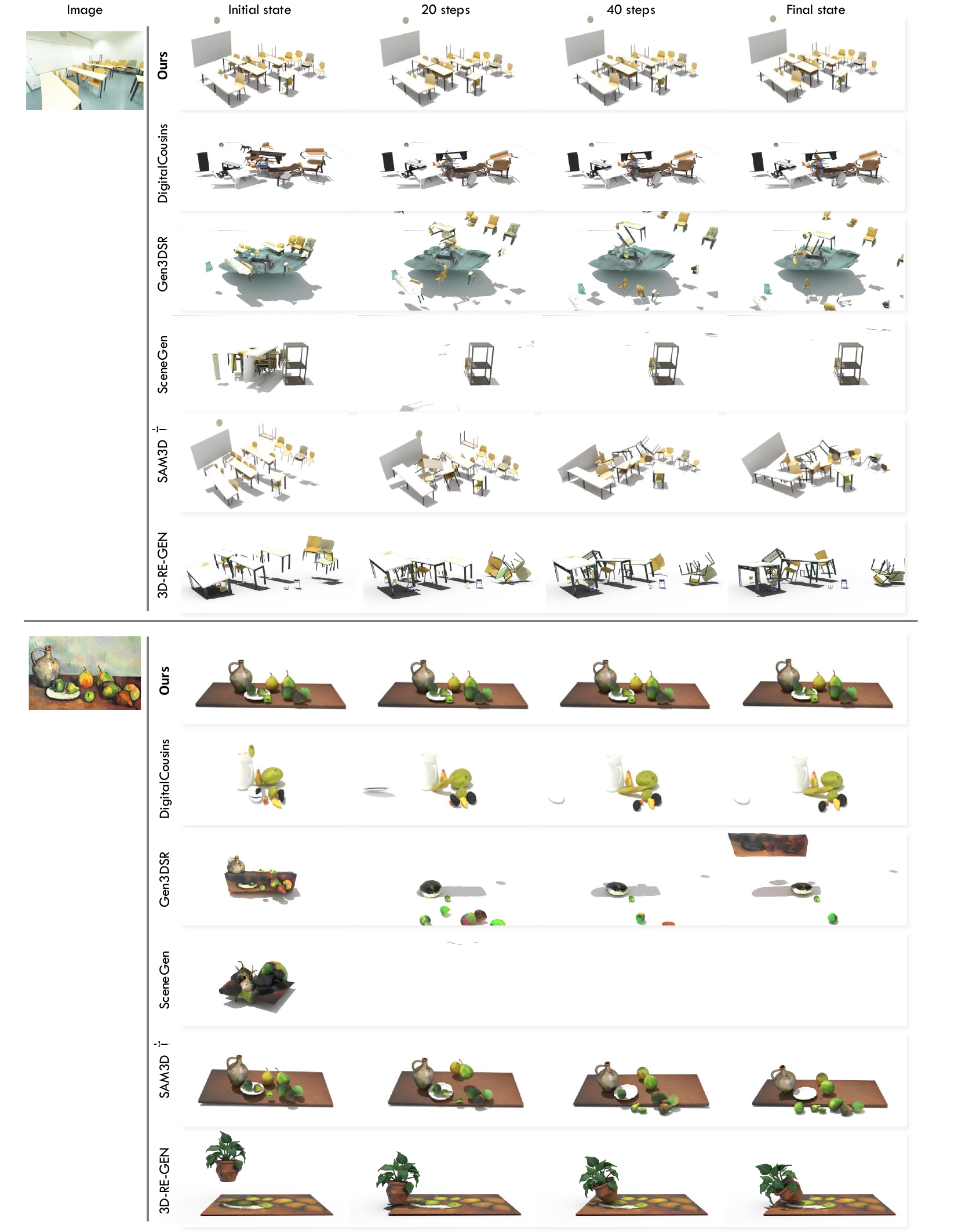}
  \caption{\textbf{Additional comparison of simulation processes of reconstructed scenes across methods.} We visualize the simulation process of scenes reconstructed by different methods in a physics simulator (Isaac Gym). Image source: (top) ScanNet++ \cite{yeshwanth2023scannetpp}; (bottom) VIGA \cite{yin2026vision}.}
  \label{fig:exp_sota2}
  \vspace{-0.8em}
\end{figure*}

\vspace{0.8em}
\noindent\textbf{Qualitative Results.}
\cref{fig:exp_sota} shows qualitative comparisons of simulation processes across different methods. We visualize reconstructed scenes in a physics simulator (Isaac Gym), showing the initial state, intermediate steps (20 and 40), and final state. In some cases, severe inter-object penetration prevents full stabilization. MIDI and DepR produce only geometry, so we omit them from the following visualizations for a fairer comparison. DigitalCousins models wall-attachment relations but may produce incorrect spatial relations (\eg chairs placed on tables), while its retrieval-based design introduces mismatched assets. SceneGen and Gen3DSR produce inaccurate layouts. SAM3D$\dagger$, combined with our Stage 1, reconstructs individual objects with relatively high fidelity, but suffers from global orientation errors, inter-object penetration, and missing wall-attachment relations, causing wall-mounted objects to fall. 3D-RE-GEN has a relatively high failure rate and fails to produce a valid result in this case, mainly due to errors in its single-forward point cloud extraction and object detection. In contrast, our method, with agentic scene analysis and physics-constrained optimization, remains stable over time while maintaining visual consistency with the input image. \cref{fig:exp_cast} compares our method with CAST. As CAST has no public code, we show only its reported results; our reconstruction is visually closer to the input image.
Please refer to Supp. \cref{supp:sec:exp} for additional qualitative results (\cref{fig:exp_sota1,fig:exp_sota2,fig:supp_exp_sota1,fig:supp_exp_sota2,fig:supp_exp_sota3,fig:supp_exp_sota4}) and video for simulation results.

\begin{table}[t]
\centering
\caption{\textbf{Ablation on scene canonicalization and global-only optimization.} SAM3D$\dagger$ denotes SAM3D with our integration.}
\vspace{-0.8em}
\label{tab:ablation_init}
\resizebox{\linewidth}{!}{
\setlength{\tabcolsep}{0.8mm}
\begin{tabular}{l cccccc}
\toprule
Ablation 
& Fail. Rate$\downarrow$ 
& Coll. Rate$\downarrow$ 
& Stable Rate$\uparrow$
& Pos. Drift$\downarrow$
& Lin. Vel.$\downarrow$
& Ang. Vel.$\downarrow$ \\
\midrule
(a) SAM3D $\dagger$ ($\sceneraw$)        &  \textbf{0.0}     &   45.1     & 12.0     &    1.111 &   1.939  & 10.246 \\
(b) Canon. ($\sceneinit$) & \textbf{0.0} & 26.3   & 45.8 & 0.767 & 1.227 & 9.724 \\
\midrule
(c) Global-only           &   10.0  & 4.3 & 89.4 & 0.116 & 0.281 & 1.903 \\
\midrule
(d) Proj-Sim (Global-only)           &   15.0  & 3.5 & 75.3 & 0.238 & 0.328 & 2.609 \\
(e) Proj-Sim (Global-local)   &   12.5  & 3.0 & 70.4 & 0.275 & 0.377 & 2.890 \\
\midrule
\textbf{Ours} ($\scene$)    &    \textbf{0.0}   & \textbf{2.7} & \textbf{94.9}     &   \textbf{0.072}  &  \textbf{0.138}  &  \textbf{0.880} \\
\bottomrule
\end{tabular}}
\end{table}

\subsection{Ablation Study}
All ablations are conducted on 3 datasets and reported as averages; geometric metrics are not available on \textit{Custom} due to missing \ac{gt}.

\vspace{0.8em}
\noindent\textbf{Scene Canonicalization.}
\cref{tab:ablation_init}(a-b) study the effect of scene canonicalization in our method. The raw output of SAM3D ($\sceneraw$) is highly unstable in physical simulation. Our scene canonicalization step ($\sceneinit$) significantly improves physical stability, reducing large drift and unstable cases. However, canonicalization alone is still insufficient to ensure full physical plausibility. By further applying physics-constrained optimization, our full model achieves the best performance. Please refer to \cref{supp:sec:expablate} for visual result of $\sceneinit$ for the same case in \cref{fig:exp_sota}. It can be observed that the canonicalized scene still remains unstable and eventually collapses during simulation. 

\begin{figure*}[t]
  \centering
  \includegraphics[width=.98\linewidth]{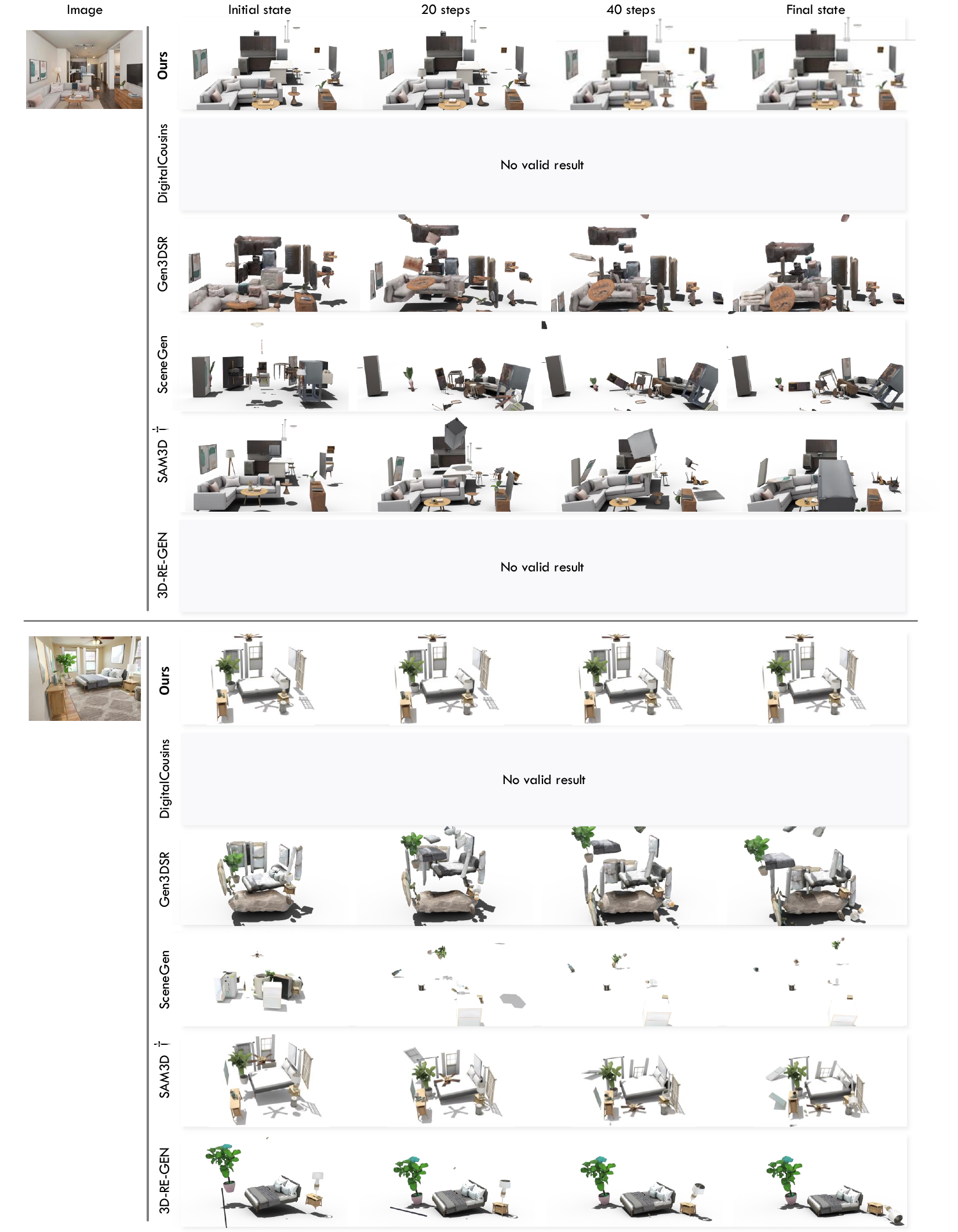}
  \caption{\textbf{Additional comparison of simulation processes of reconstructed scenes across methods on our custom data.} We visualize the simulation process of scenes reconstructed by different methods in a physics simulator (Isaac Gym). DigitalCousins \cite{dai2024acdc} fails to produce any valid results. Our method remains stable while others exhibit noticeable instability. Image source: Internet.}
  \label{fig:exp_sota1}
  \vspace{-0.8em}
\end{figure*}

\vspace{0.8em}
\noindent\textbf{Global-only Optimization.}
We ablate the effect of the local group optimization stage in our physics-constrained optimization by removing the scene-tree guided step and instead optimizing all objects jointly in a single global optimization. \cref{tab:ablation_init}(c) reports the results. We observe that global-only optimization fails to produce valid results in approximately 10\% of cases. This is mainly because some scenes contain a large number of objects, making joint optimization difficult and unstable. As a result, several cases completely fail to converge, \eg for the top case of \cref{fig:exp_sota1}, the global-only model produces no valid output. Therefore, we report metrics on successful cases only, which may be slightly optimistic as harder cases are excluded due to failure. Even so, it still underperforms our full scene-tree-guided method. 

\vspace{0.8em}
\noindent\textbf{Projection-based Optimization.}
We further replace CEM with a projection-based alternative following \cite{pfaff2025_steerable_scene_generation}, which resolves penetration via nonlinear constrained optimization. We evaluate two variants: \cref{tab:ablation_init}(d) projects all objects, while (e) applies our scene-tree-guided strategy. Both are less reliable, failing on over 10\% of scenes, as they optimize only translations with fixed rotations and may not converge on cluttered scenes. Variant (e) reduces failures but degrades stability because only group-root geometry is optimized, leaving child objects out of joint projection and resulting in coarser inter-group contact resolution. While projection is faster (2.5 min \vs 10.4 min/scene), it sacrifices success rate and stability.

\begin{figure}[t]
  \centering
  \includegraphics[width=\linewidth]{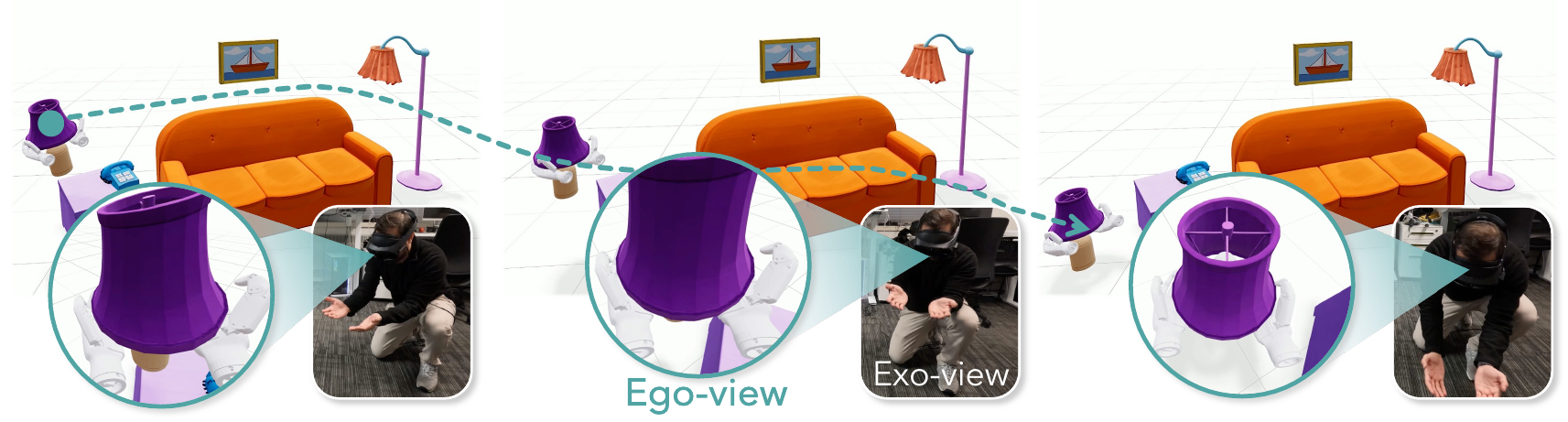}
 \caption{\textbf{VR interaction system demo.} We implement a VR interaction system that enables users to naturally interact with reconstructed scenes in real time using a VR headset.}
  \label{fig:exp_vr}
\end{figure}

\subsection{Real-world VR Interaction}
To demonstrate the physical plausibility of our reconstructions, we implement a VR interaction system that allows users to naturally interact with the virtual objects in real time (30 FPS) using a VR headset (Meta Quest Pro). Hand motions are tracked and mapped to a dexterous robotic hand in Isaac Gym for interaction under simulation. The simulation is rendered back to the headset. \cref{fig:exp_vr} shows an example of the VR interaction process. This further highlights the potential of our system for content creation, where users can naturally manipulate and rearrange reconstructed environments. See Supp. \cref{supp:sec:vr} and \ourpagetext for details and demos.

\section{Conclusion}
\label{sec:conclusion}
We present \ourmethod, a framework that transforms casual images into simulation-ready 3D digital assets by integrating physical scene understanding with physics-constrained refinement, pushing beyond prior work that focuses primarily on visual plausibility. We introduce an agentic physical scene understanding module that infers object physical states and inter-object relationships as structural priors for physics-constrained optimization, enabling correction of physical errors while preserving consistency with the input image. Experiments show that our method outperforms \ac{sota} approaches in both reconstruction quality and physical plausibility. We further demonstrate the reconstructed scenes via VR-based human–object interaction, highlighting their potential for immersive applications.

\vspace{0.8em}
\noindent\textbf{Limitations.}
Our method relies on the robustness of VLMs for physical scene understanding and may fail in challenging cases. The reconstructed scenes may also deviate from the input due to object reconstruction errors or physics-induced displacements. Stronger image-to-3D models can be readily integrated to improve fidelity. Currently, we focus on rigid objects and do not explicitly model deformable or non-rigid objects, which we leave for future work.

\section*{Acknowledgment}
The authors would like to thank Jiawei Liu for his help with issues related to SAM3D, and Yuxuan Kuang, Yufei Wang, and Maxwell Jones for their insightful discussions.

\bibliographystyle{ieeenat_fullname}
\bibliography{reference}

\clearpage
\appendix

This supplementary material provides additional details on the method (\cref{supp:sec:method}) and experimental setup (\cref{supp:sec:implement}), along with further experimental results and extended visualizations (\cref{supp:sec:exp}).

\section{Method Details}
\label{supp:sec:method}
\subsection{Scene Tree Construction}
\label{supp:sec:tree}
\noindent\textbf{Open-vocabulary Object List Analysis.}
To enable the \ac{vlm} to generate a structured open-vocabulary object list $\objlist$, we prompt it to detect salient objects. When multiple similar objects are present, we require additional disambiguation using location or color attributes. Besides, we manually include the floor as a canonical ground-plane reference in the object list. The full prompt is provided in our \ourrepotext. \\

\noindent\textbf{Agentic Instance Segmentation.}
The agentic instance segmentation system consists of a segmentation agent $\agentseg$ and a verifier agent $\agentv$. The segmentation agent $\agentseg$ takes the object list $\objlist$ as input and performs segmentation via tool calling, invoking a local SAM3 \cite{carion2026sam} model. Specifically, $\agentseg$ first describes each object in $\objlist$ and then obtains the corresponding mask predictions. The resulting masks are overlaid onto the input image and passed to $\agentv$ for verification.

The verifier agent $\agentv$ evaluates whether the overlaid masks are correct. It iteratively interacts with $\agentseg$, refining or rejecting incorrect masks, until either the mask is verified as correct by $\agentv$, or the object is determined to be absent after 10 iterations.

\begin{figure}[h]
\centering
\includegraphics[width=\linewidth]{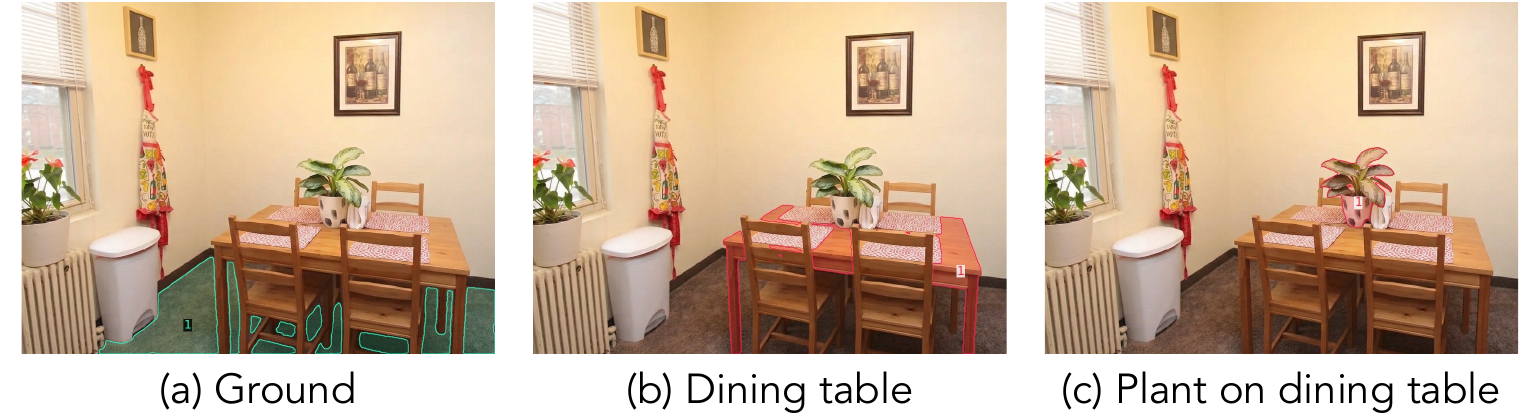}
\caption{\textbf{Examples of overlaid object masks.} The images are provided to the verifier agent $\agentv$ to determine mask correctness and also assist in constructing the scene tree.}
\label{supp:fig:vlm}
\end{figure}

An example of the overlaid masks is shown in \cref{supp:fig:vlm}, which illustrates three objects: (a) ground, (b) dining table, and (c) plant on the dining table. The overlaid masks serve as visual inputs to the verifier agent for mask validation and also for assisting scene graph construction in the subsequent step.

\noindent\textbf{Scene Tree Construction via Spatial Reasoning.}
Our scene tree contains four canonical roots: \textit{ground}, \textit{wall}, \textit{ceiling}, and \textit{ground-wall}. We aim to assign each object its corresponding parent under a gravity-aware perspective, and further infer the spatial relation between the object and its parent (\ie, on, inside, hanging, or attached), as well as whether the object is \textit{movable} or \textit{fixed}.

To this end, we prompt a \ac{vlm} to perform this reasoning task, leveraging the overlaid masks from the previous step to better localize objects. The composite root, \ie ground-wall, is introduced to handle cases where objects are simultaneously supported by the ground and constrained by the wall, such as the ``white radiator'' shown in \cref{supp:fig:vlm}. The hanging and attached relations primarily apply to wall- or ceiling-mounted objects; for example, curtains are hanging from a rod, paintings are attached to walls, and ceiling lamps are attached to the ceiling.

As described in \cref{sec:opt}, objects whose parents are walls or ceilings are excluded from local and global optimization. Instead, they are placed (see \cref{supp:sec:phys_opt} for details) after optimizing the remaining scene objects and are fixed during simulation.

We tested different prompt variations, minor prompt changes have minimal impact. The final full prompt for this reasoning stage is provided in our \ourrepotext.

\begin{algorithm}[t]
\caption{Physics-Constrained Optimization}
\label{alg:phys_opt}
\small
\DontPrintSemicolon
\KwIn{Canonicalized scene $\sceneinit$, scene tree $\scenetree$.}
\KwOut{Optimized scene $\scene$.}

$\mathcal{G} =$ local groups in $\scenetree$ (non-root nodes with $\geq 1$ child)\;
$\mathcal{R} = \{\text{root of }\groupidx \in \mathcal{G}\} \cup \{i \in \scenetree : \mathrm{parent}(i) \in \{\text{ground},\,\text{ground-wall}\}\}$\;

\BlankLine
\tcc{(a) Local group optimization}
\ForEach{$\groupidx \in \mathcal{G}$ \rm{(post-order)}}{
  $\groupinit_\groupidx = \{(\rotinit_i, \transinit_i)\}_{i \in \groupidx}$ from $\sceneinit$\;
  $\groupbest_\groupidx = \textsc{CEM}(\groupinit_\groupidx)$ \tcp*{\cref{alg:cem}}
  $\{(\rotinit_i, \transinit_i)\}_{i \in \groupidx} \gets \groupbest_\groupidx$ in $\sceneinit$ \tcp*{update group layout}
}

\BlankLine
\tcc{(b) Global layout optimization}
$\groupinit_\mathcal{R} = \{(\rotinit_r, \transinit_r)\}_{r \in \mathcal{R}}$ from $\sceneinit$\;
$\groupbest_\mathcal{R} = \textsc{CEM}(\groupinit_\mathcal{R})$

\BlankLine
\tcc{(c) Wall and ceiling object placement}
$\mathcal{W} = \{i \in \scenetree : \mathrm{parent}(i) \in \{\text{wall},\,\text{ceiling}\}\}$\;
$\groupbest_\mathcal{W} = \textsc{PlaceWallCeiling}(\mathcal{W},\, \groupbest_\mathcal{R})$ \tcp*{\cref{supp:sec:placewall}}

\BlankLine
$\scene = \groupbest_\mathcal{R} \,\cup\, \{\groupbest_\groupidx\}_{\groupidx \in \mathcal{G}} \,\cup\, \groupbest_\mathcal{W}$ \tcp*{Optimized scene}
\Return $\scene$\;
\end{algorithm}

\begin{algorithm}[t]
\caption{$\textsc{CEM}$ Optimization}
\label{alg:cem}
\DontPrintSemicolon
\KwIn{Initial layout $\groupinit$.}
\KwOut{Best layout $\groupbest$.}
Initialize $(\boldsymbol{\mu}_1, \boldsymbol{\Sigma}_1)$\;
\For{$t = 1, \ldots, T$}{
  Sample $\Delta\group_t^{(k)} \sim \mathcal{N}(\boldsymbol{\mu}_t, \boldsymbol{\Sigma}_t)$ for $k = 1, \ldots, K$\;
  $\groupk_t = \groupinit \oplus \Delta\group_t^{(k)}$  \tcp*{Candidate layout}
  Evaluate $E(\groupk_t)$ by forward simulation in Isaac Gym\;
  $\elite_t = \{k : E(\groupk_t) \text{ is among the } \lceil \rho K \rceil \text{ lowest}\}$\;
  $\boldsymbol{\mu}_{t+1} \gets \dfrac{1}{|\elite_t|} \sum_{k \in \elite_t} \Delta\group_t^{(k)}$,\,
  $\boldsymbol{\Sigma}_{t+1} \gets \dfrac{1}{|\elite_t|} \sum_{k \in \elite_t} \bigl(\Delta\group_t^{(k)} - \boldsymbol{\mu}_{t+1}\bigr)^2$\;
}
$\groupbest = \arg\min_{t,\,k}\, E(\groupk_t)$\;
\Return $\groupbest$\;
\end{algorithm}

\subsection{Physics-Constrained Optimization}
\label{supp:sec:phys_opt}
\cref{alg:phys_opt} summarizes the full physics-constrained optimization procedure described in \cref{sec:opt}. Stage (a) traverses the scene tree $\scenetree$ in post-order and runs CEM optimization independently on each local group. Stage (b) treats each optimized group as a rigid unit and runs a second CEM pass jointly over group roots and ground- or ground-wall-supported objects. Both stages share the CEM subroutine (see \cref{alg:cem}), which evaluates each candidate by forward simulation in Isaac Gym~\cite{makoviychuk2021isaac} and updates the sampling Gaussian by moment matching over the elite set $\elite$. Final wall- and ceiling-attached objects are placed by the heuristic detailed as follows. \\

\noindent\textbf{Wall and Ceiling Object Placement.}
\label{supp:sec:placewall}
Objects whose scene-tree parent is \textit{wall} or \textit{ceiling} are initially placed at their original positions from $\sceneinit$. 
If interpenetration with the scene is detected, we first estimate a simple room structure by fitting three walls (back, left, and right) from the axis-aligned bounding box of all already settled objects, where each wall is represented as a planar mesh. Each wall/ceiling object is then assigned to its nearest fitted surface. After assignment, objects are translated along the normal direction of their associated surface until all intersections with the already optimized scene 
$\groupbest_\mathcal{R} \cup \{\groupbest_g\}_{g \in \mathcal{G}}$ are resolved, ensuring a collision-free placement.

\begin{figure*}[t]
  \centering
  \includegraphics[width=.72\linewidth]{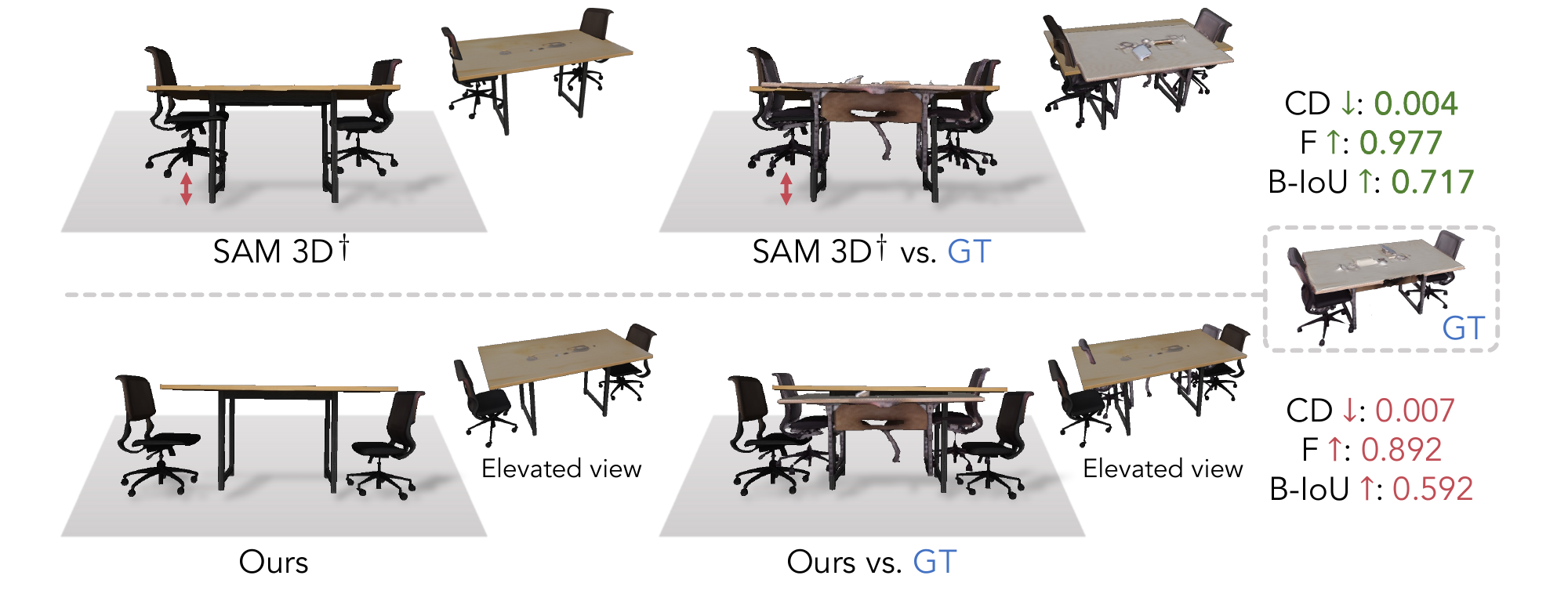}
  \caption{\textbf{A typical case illustrating the limitation of geometric metrics.} SAM3D $\dagger$ produces physically implausible reconstructions (\eg, floating objects and inter-penetration), which can still achieve better geometric scores after ICP alignment. In contrast, our method enforces physical plausibility, leading to worse geometric metrics despite more stable results.}
  \label{supp:fig:cd_error}
\end{figure*}

\section{Experimental Setup}
\label{supp:sec:implement}

\subsection{Evaluation Setup}
\noindent\textbf{Baselines.} For retrieval-based Digital Cousins \cite{dai2024acdc}, each input scene produces three retrieved candidates. We report quantitative results by averaging over all three candidates, while visualization results are shown using the first candidate.

Note that SAM3D is designed for single-object reconstruction and thus takes an individual object as input. 
To adapt it to scene-level reconstruction, we first apply our Stage 1 pipeline to obtain instance-level masks. 
These masks are then used to independently reconstruct each object with SAM3D, and the resulting outputs are composed into a full scene, denoted as $\sceneraw$ (see \cref{sec:init}). 
We mark this modification with a $\dagger$ in the following tables and figures. \\

\noindent\textbf{Dataset.} Following \cite{ni2024phyrecon,yu2022monosdf}, we evaluate on 10 scenes from Replica~\cite{replica19arxiv} and 12 scenes from ScanNet++~\cite{yeshwanth2023scannetpp}. We use per-object GT meshes provided by prior work \cite{ni2024phyrecon}. In addition, the \textit{custom} evaluation set consists of 25 images collected from the Internet and prior work \cite{xia2026sage,sautter20253dregen,marble2025worldlabs,yin2026vision} (\ie, VIGA, SAGE, 3D-RE-GEN). The dataset covers a diverse range of scene types, including real-world images, synthesized renderings, cartoon-style images, and Gaussian-rendered scenes (from World Labs Marble \cite{marble2025worldlabs}). Each scene contains a varying number of objects, ranging from 4 to 30. \\

\label{supp:sec:metric}
\noindent\textbf{Metrics.}
For physical metrics, we compute the collision rate using trimesh \cite{trimesh2019} by detecting mesh intersections among reconstructed 3D objects. The reconstructed scene is then imported into Isaac Gym for physics-based simulation. An object is considered unstable if its relative translation exceeds 0.1\,m or its rotation exceeds 0.1\,rad after $L=60$ simulation steps. We report position drift as the displacement between the initial placement and the final settled state. During simulation, we further record the average peak linear and angular velocities of all objects.

For geometric metrics, \ac{cd} measures the bidirectional nearest-neighbor distance between reconstructed and \ac{gt} point sets. F-score@0.05 (F-Score) evaluates correspondence accuracy under a 0.05 distance threshold. B-IoU measures the intersection-over-union between axis-aligned 3D bounding boxes of reconstructed and \ac{gt} objects, ranging from 0 (no overlap) to 1 (perfect overlap).

\subsection{Implementation Details}
\label{supp:sec:expdetail}
For instance segmentation, the agentic loop calls SAM3~\cite{carion2026sam} as a tool, with a maximum of 10 iterations for $\agentv$ to verify.

For physics-constrained optimization, the initial Gaussian has zero mean and per-DoF standard deviations $\boldsymbol{\sigma}_0^{\text{trans}} = (0.05, 0.005, 0.05)$\,m and $\boldsymbol{\sigma}_0^{\text{rot}} = (0.005, 0.05, 0.005)$\,rad, reflecting the prior that horizontal displacement and pitch require larger search ranges than vertical displacement and roll/yaw after canonicalization. 

For physics simulation, we use Isaac Gym~\cite{makoviychuk2021isaac} with the GPU PhysX backend. 
We set 6 position and 1 velocity iterations, a contact offset of 0.01\,m, and a maximum depenetration velocity of 5\,m/s. Simulation runs with a timestep of $1/60$\,s and 2 substeps, and gravity is set to $-9.8$\,m/s$^2$ along the $-Y$ direction. The ground plane uses static and dynamic friction of 1.0 with zero restitution.
For object assets, we apply V-HACD convex decomposition with a voxel resolution of $10^5$. The maximum number of convex hulls is set to 8 for near-convex furniture, 32 for organic shapes (\eg, plants and curtains), and 16 for other categories. The category of each object is obtained from the object descriptions with semantic keywords extracted by the \ac{vlm} in Stage 1, which we manually map to the above categories. The hull counts are kept low to balance simulation accuracy and runtime. We further set linear and angular damping to 0.3, and override center of mass and inertia.
Each candidate is simulated for $\simstep=60$ forward steps. Velocity energy is measured at step $\tau = 15$ as an early indicator of instability.

\section{Additional Experimental Results}
\label{supp:sec:exp}
This section provides additional results and analyses, including comparisons with \ac{sota} methods (\cref{supp:sec:expresults}), geometric metric analysis, extended visualizations, efficiency and API cost evaluation, extensive ablation studies (\cref{supp:sec:expablate}), as well as discussions of failure cases and limitations (\cref{supp:sec:expdiscuss}).

\begin{table}[t]
\centering
\caption{\textbf{Mean and standard deviation across scenes for each dataset using our method.} R, S, C represent Replica, ScanNet++, and our Custom datasets, respectively.}
\label{tab:supp:per_dataset_std}
\resizebox{\linewidth}{!}{
\setlength{\tabcolsep}{1.5mm}
\begin{tabular}{llccccccccc}
\toprule
\multirow{3}{*}{Data} & \multirow{2}{*}{} & \multicolumn{5}{c}{Physical Metrics} & \multicolumn{3}{c}{Geometric Metrics} \\
\cmidrule(lr){3-7} \cmidrule(lr){8-10}
&
& \shortstack{Coll.\\Rate (\%)}$\downarrow$
& \shortstack{Stable\\Rate (\%)}$\uparrow$
& \shortstack{Pos.\\Drift (m)}$\downarrow$
& \shortstack{Lin. Vel.\\(m/s)}$\downarrow$
& \shortstack{Ang. Vel.\\(rad/s)}$\downarrow$
& CD$\downarrow$
& F-Score$\uparrow$
& B-IoU$\uparrow$ \\
\midrule

\multirow{2}{*}{R}
& Mean & 6.7 & 96.2 & 0.037 & 0.169 & 0.539 & 0.011 & 0.868 & 0.38 \\
& Std  & 10.9 & 6.0 & 0.084 & 0.210 & 0.493 & 0.009 & 0.104 & 0.17 \\

\midrule
\multirow{2}{*}{S}
& Mean & 5.9 & 93.6 & 0.080 & 0.159 & 1.039 & 0.019 & 0.807 & 0.20 \\
& Std  & 9.8 & 7.1 & 0.091 & 0.121 & 1.287 & 0.013 & 0.093 & 0.11 \\

\midrule
\multirow{2}{*}{C}
& Mean & 1.2 & 95.5 & 0.017 & 0.140 & 0.468 & --- & --- & --- \\
& Std  & 4.5 & 5.1 & 0.107 & 0.105 & 0.827 & --- & --- & --- \\

\bottomrule
\end{tabular}}
\end{table}

\subsection{Comparison with SOTAs}
\label{supp:sec:expresults}
\noindent\textbf{Per-Scene Statistics.}
To assess metric consistency across scenes, we report the mean and standard deviation of each metric within each dataset in \cref{tab:supp:per_dataset_std}. Our method shows consistently strong physical stability across scenes and datasets, demonstrating robust performance across diverse scenarios. \\

\noindent\textbf{Case Study of Geometric Metrics.}
As analyzed in \cref{sec:exp}, our method achieves the best trade-off between physical stability and fidelity to the input. On geometric metrics, our results are comparable to SAM3D~\cite{chen2025sam3d}, with a better \ac{cd} but slightly worse F-Score and B-IoU. These metrics measure geometric alignment with \ac{gt} after ICP alignment, but do not account for scene-level physical stability. Thus, they cannot fully capture the physical improvements introduced by our optimization. The slightly lower F-Score and B-IoU are partly due to the small object displacements introduced when resolving physical errors.

To better illustrate this, we visualize a typical case in \cref{supp:fig:cd_error}. We show results from SAM3D (top) and our method (bottom), along with the ICP-aligned \ac{gt} (second column). SAM3D exhibits severe interpenetration and floating objects, while our method separates the chairs from the table and places the objects stably on the ground. This example highlights that geometric metrics alone do not fully reflect scene-level physical stability. Some geometric errors from the underlying image-to-3D model can also lead to larger displacements during physical correction, which remains a limitation that could be alleviated by stronger image-to-3D models in the future. Therefore, we include these metrics only as a proof-of-concept comparison. \\

\noindent\textbf{Object Detection Accuracy.}
To evaluate our agentic scene analysis design, \ie the agentic instance segmentation module, we further evaluate the accuracy of detecting open-vocabulary objects on ScanNet++~\cite{yeshwanth2023scannetpp} in \cref{supp:tab:detection}, which provides \ac{gt} instance masks. For each \ac{gt} object, we match it to the predicted mask with the highest IoU and report Recall@0.5, mean IoU (mIoU) over matched pairs, and Miss Rate, \ie the percentage of \ac{gt} objects that are not detected. Prior methods usually rely on a single-pass, fixed-threshold object detector (\eg, GroundedSAM \cite{ren2024grounded}) without any verification mechanism, so they remain brittle to missed detections. Even though MIDI, DepR, and 3D-RE-GEN are all built upon GroundedSAM, their differences mainly stem from the different underlying object lists and model sizes adopted in their respective official implementations. In contrast, our agentic instance segmentation module automatically verifies each candidate mask through an iterative segmentation--verification loop between $\agentseg$ and $\agentv$, refining or rejecting uncertain predictions before acceptance. As shown in \cref{supp:tab:detection}, this closed-loop agentic design substantially improves detection reliability, achieving the best performance among all compared methods. \\

\begin{table}[t]
\centering
\small
\caption{\textbf{Object detection accuracy on ScanNet++.} We compare our agentic instance segmentation module against the detection components used in prior methods.}
\label{supp:tab:detection}
\resizebox{\linewidth}{!}{
\setlength{\tabcolsep}{2mm}
\begin{tabular}{llccc}
\toprule
Method & Detector & Recall@0.5$\uparrow$ & mIoU$\uparrow$ & Miss Rate$\downarrow$ \\
\midrule
MIDI \cite{huang2025midi} & GroundedSAM & 0.67 & 0.66 & 0.28 \\
DepR \cite{Zhao_2025_ICCV_DepR} & GroundedSAM & 0.79 & 0.73 & 0.21 \\
Gen3DSR \cite{Ardelean2025Gen3DSR} & CropFormer & 0.79 & 0.74 & 0.21 \\
3D-RE-GEN \cite{sautter20253dregen} & GroundedSAM & 0.85 & 0.69 & 0.15 \\
\midrule
\textbf{Ours} & \textbf{Agentic} & \textbf{0.90} & \textbf{0.83} & \textbf{0.05} \\
\bottomrule
\end{tabular}}
\end{table}

\begin{figure}[t]
  \centering
  \includegraphics[width=\linewidth]{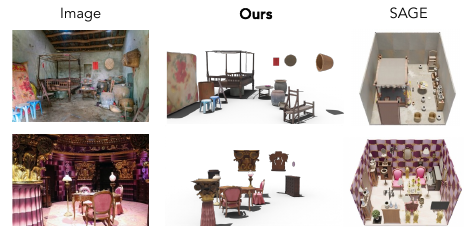}
  \caption{\textbf{Comparison with scene generation method SAGE.} Fully agentic method SAGE \cite{xia2026sage} lacks controllability and may produce inconsistent reconstructions, while our method yields more faithful results while remaining simulation-ready.}
  \label{supp:fig:exp_sage}
\end{figure}

\noindent\textbf{Comparison with Scene Generation Method.}
We further compare our method with recent fully agentic scene generation approaches such as SAGE \cite{xia2026sage} in \cref{supp:fig:exp_sage}. While these methods can generate simulation-ready scenes, their fully agent-driven pipelines often lack controllability, leading to heuristic, template-like results and inconsistencies when applied to real-world reconstruction. In contrast, our method targets the more constrained task of single-image reconstruction, where the generated scene should faithfully match the input observation. As can be seen, our method achieves significantly more reliable and faithful reconstructions from casual images. This improvement can be attributed to our three-stage agent-assisted reconstruction and physics-constrained optimization, which enhance controllability over the generation process while still producing simulation-ready scenes. \\

\begin{figure*}[t]
  \centering
  \includegraphics[width=.9\linewidth]{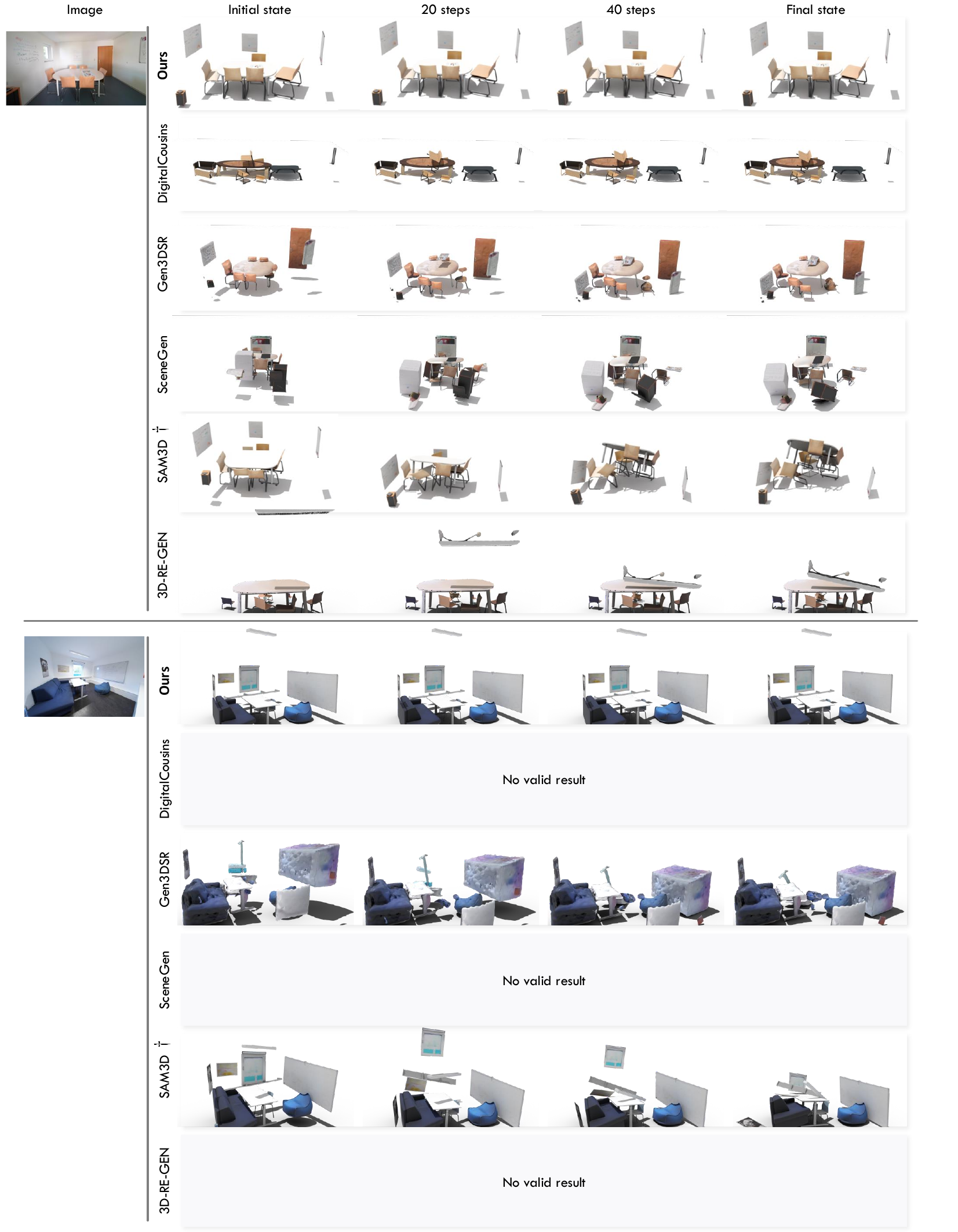}
  \caption{\textbf{Additional qualitative comparison of simulation processes.} We visualize the simulation process of scenes reconstructed by different methods in a physics simulator (Isaac Gym). DigitalCousins~\cite{dai2024acdc} retrieves objects that significantly deviate from the input scene with incorrect spatial layout. SceneGen and SAM3D$\dagger$ suffer from severe interpenetration, leading to unstable simulations. Image source: ScanNet++~\cite{yeshwanth2023scannetpp}.}
  \label{fig:supp_exp_sota1}
  \vspace{-0.8em}
\end{figure*}

\begin{figure*}[t]
  \centering
  \includegraphics[width=.9\linewidth]{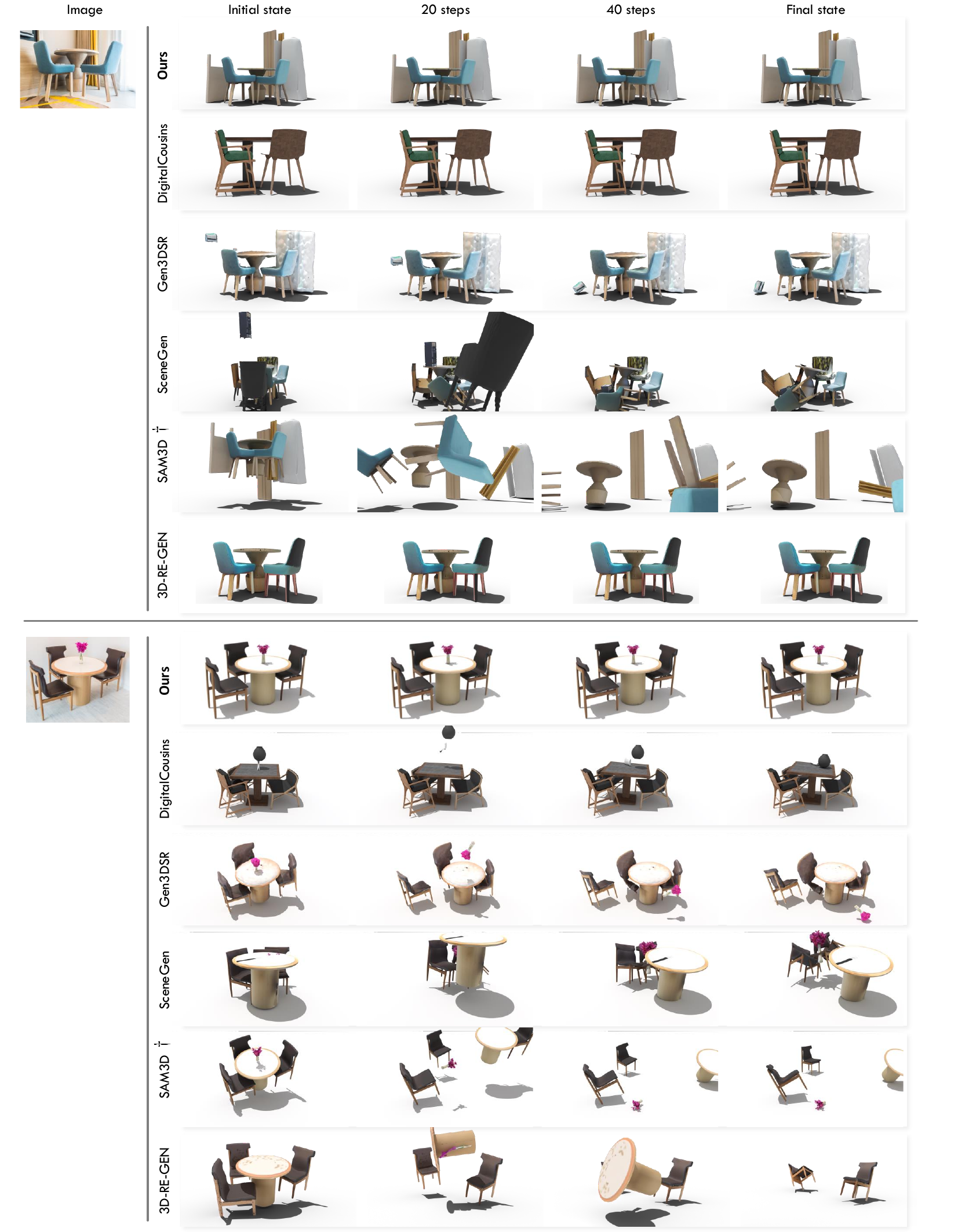}
  \caption{\textbf{Additional qualitative comparison of simulation processes.} We visualize the simulation process of scenes reconstructed by different methods in a physics simulator (Isaac Gym). SceneGen and SAM3D$\dagger$ suffer from severe interpenetration, leading to unstable simulations. Image source: 3D-RE-GEN \cite{sautter20253dregen}. }
  \label{fig:supp_exp_sota2}
  \vspace{-0.8em}
\end{figure*}

\begin{figure*}[t]
  \centering
  \includegraphics[width=.9\linewidth]{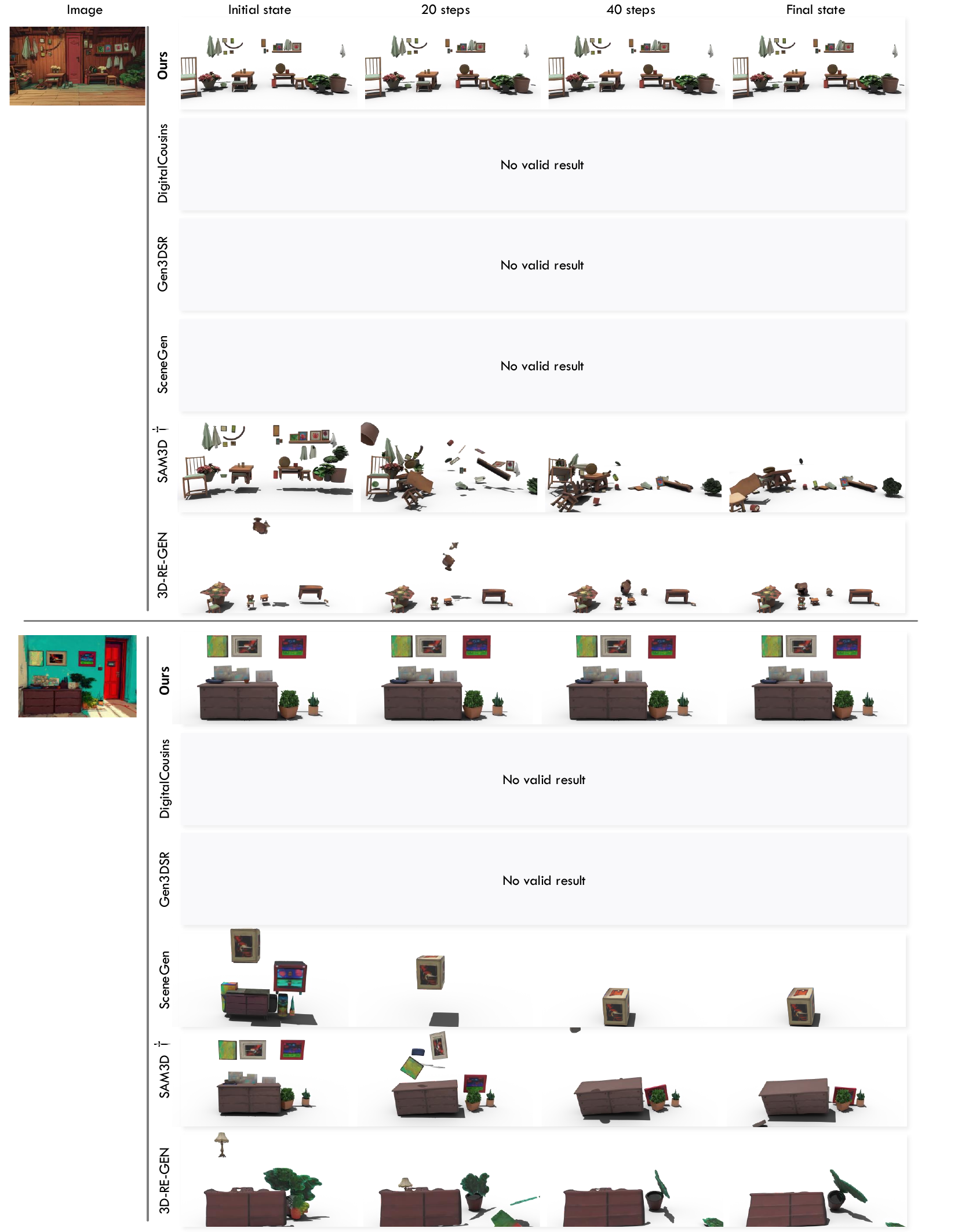}
  \caption{\textbf{Additional qualitative comparison of simulation processes.} We visualize the simulation process of scenes reconstructed by different methods in a physics simulator (Isaac Gym). Our method supports Gaussian-rendered scenes from World Labs Marble \cite{marble2025worldlabs} as input and produces the most stable results, while DigitalCousins and Gen3DSR fail to yield valid outputs. Image source: World Labs Marble \cite{marble2025worldlabs}. }
  \label{fig:supp_exp_sota3}
  \vspace{-0.8em}
\end{figure*}

\begin{figure*}[t]
  \centering
  \includegraphics[width=.9\linewidth]{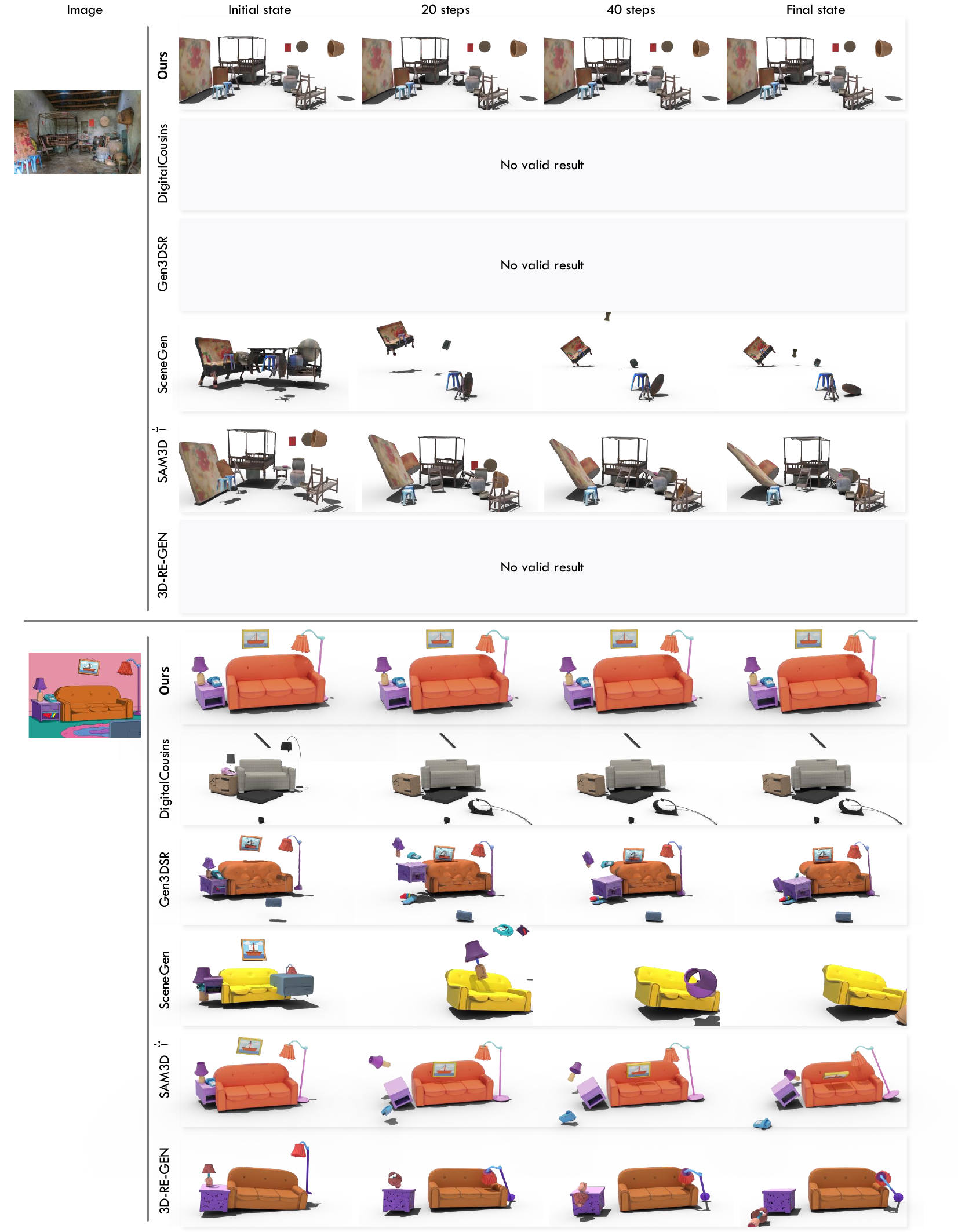}
  \caption{\textbf{Additional qualitative comparison of simulation processes.} We visualize the simulation process of scenes reconstructed by different methods in a physics simulator (Isaac Gym). Our method remains robust on synthetic input, while other methods either fail to produce valid results or exhibit severe instability. Image source: (top) SAGE \cite{xia2026sage}; (bottom) Internet.}
  \label{fig:supp_exp_sota4}
  \vspace{-0.8em}
\end{figure*}

\noindent\textbf{Additional Qualitative Results.}
We present extensive qualitative comparisons of simulation processes across different methods in \cref{fig:supp_exp_sota1,fig:supp_exp_sota2,fig:supp_exp_sota3,fig:supp_exp_sota4}. 

As shown in \cref{fig:supp_exp_sota1}, on ScanNet++ scenes, retrieval-based Digital Cousins~\cite{dai2024acdc} often produces layouts that significantly deviate from the input scene. 
In contrast, SceneGen~\cite{meng2026scenegen} and SAM3D$\dagger$ both exhibit severe interpenetration, resulting in unstable physical simulations. In contrast, our method obtains physically stable results with correct spatial layout. \cref{fig:supp_exp_sota2} further illustrates similar failure patterns of the baselines, which suffer from interpenetration and unstable simulations, often causing objects to drift out of view during rollout, while our method remains robust.

In addition, our method generalizes well to diverse input styles. As shown in \cref{fig:supp_exp_sota3}, we can directly handle Gaussian-rendered scenes from World Labs Marble~\cite{marble2025worldlabs}, producing stable and physically plausible reconstructions, whereas baselines such as Digital Cousins~\cite{dai2024acdc} and Gen3DSR~\cite{Ardelean2025Gen3DSR} fail to produce valid outputs. \cref{fig:supp_exp_sota4} further demonstrates robustness on synthetic and cartoon-style inputs, respectively, where our method consistently maintains physically valid layouts, while competing approaches either collapse or produce unstable reconstructions. 

Please also refer to the supplementary video and project website for better visualization of the simulation results. \\

\begin{table}[t]
\centering
\small
\caption{\textbf{Inference Time and API Cost Comparison.} We report per-scene and per-object inference time for all methods. -- means not applicable.}
\label{tab:cost_time}
\resizebox{\linewidth}{!}{
\setlength{\tabcolsep}{2.5mm}
\begin{tabular}{lccc}
\toprule
{Method} & {\textit{min}/Scene} & {\textit{min}/Obj} & {Cost/Scene} \\
\midrule
DigitalCousins \cite{dai2024acdc} & 32.5  & 2.5  & \$1.25 \\
Gen3DSR \cite{Ardelean2025Gen3DSR} & 39.3  & 2.1  & -- \\
SceneGen \cite{meng2026scenegen} & 27.1  & 2.1  & -- \\
SAM3D $\dagger$ \cite{chen2025sam3d} & 10.6  & 0.8 & \$0.22  \\
\midrule
\textbf{Ours (Total)} & 25.8 & 2.0 & \$0.47  \\
\midrule
Ours (Stage 1) & 10.6  & 0.8 & \$0.22 \\
Ours (Stage 2) & 4.8  & 0.3 &  \$0.25 \\
Ours (Stage 3) & 10.4  & 0.9 & -- \\
\bottomrule
\end{tabular}}
\end{table}

\begin{figure*}[t]
  \centering
  \includegraphics[width=0.95\linewidth]{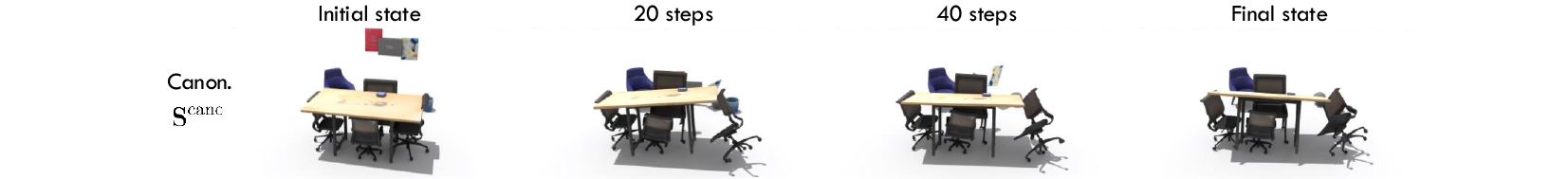}
  \caption{\textbf{Effect of scene canonicalization.} We visualize the simulation of the canonicalized scene ($\sceneinit$) for the case in \cref{fig:exp_sota}, which still remains unstable.}
  \label{supp:fig:exp_ablation_canon}
\end{figure*}

\begin{figure*}[t]
  \centering
  \includegraphics[width=.93\linewidth]{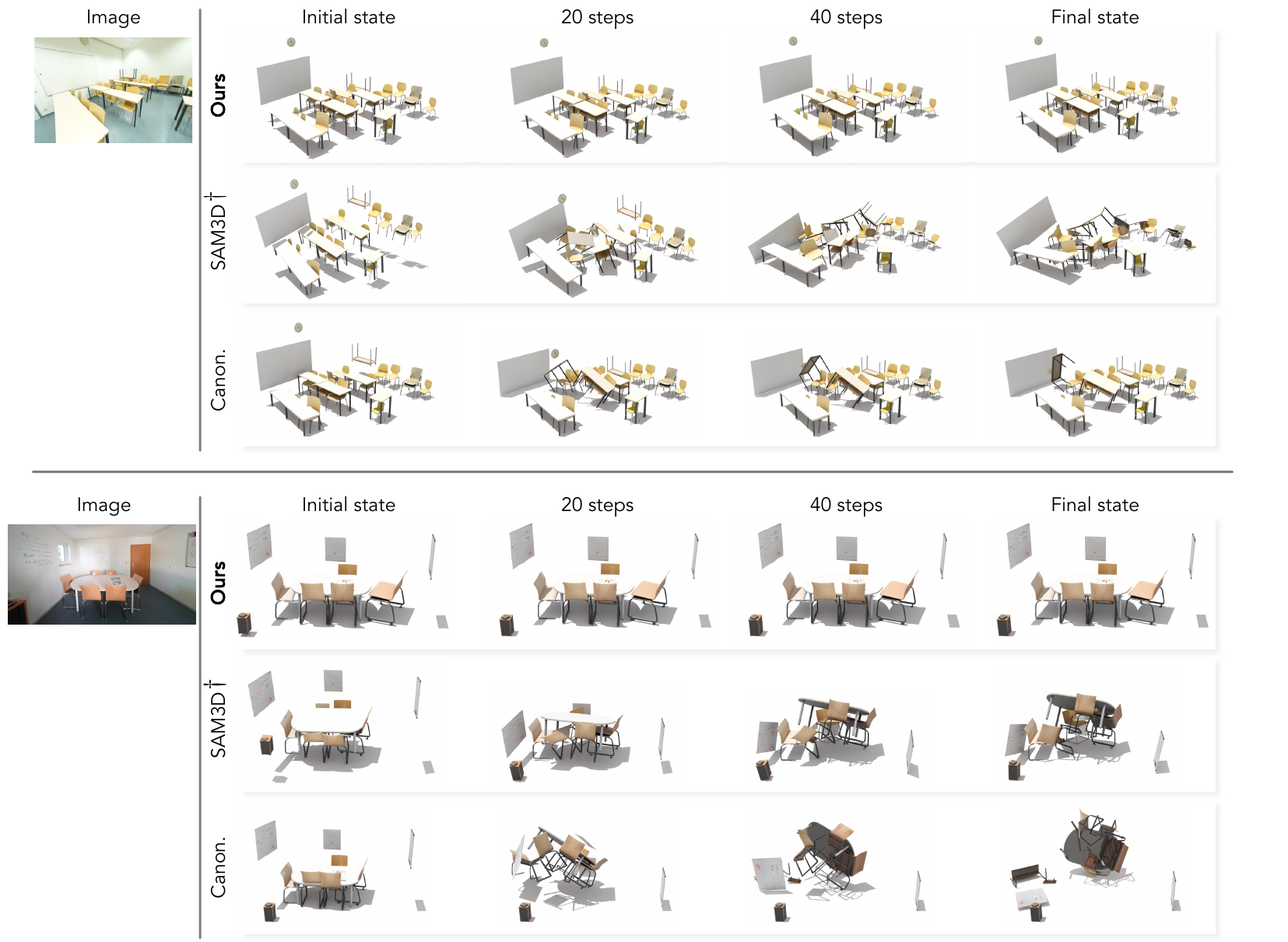}
  \caption{\textbf{Effect of scene canonicalization.} We visualize the simulation of SAM3D reconstructed scene ($\sceneraw$), the canonicalized scene (Canon. $\sceneinit$), which still remains unstable.}
  \label{supp:fig:scene_upfix}
\end{figure*}

\noindent\textbf{Inference Time and API Cost.}
We compare inference time on an NVIDIA TITAN RTX and API cost across methods in \cref{tab:cost_time}.
Among baselines, Gen3DSR~\cite{Ardelean2025Gen3DSR} is the most time-consuming. SAM3D$\dagger$ leverages our Stage~1 pipeline for scene-level reconstruction, and thus shares the same runtime and cost as our Stage~1. 
For completeness, we report the runtime and cost of each stage in our pipeline. In total, our method requires 25.8 minutes per scene, making it comparable to or faster than existing baselines. Note that the reported runtime for both DigitalCousins \cite{dai2024acdc} and our method includes API calls as well as the waiting time incurred by API requests.

In terms of cost, following the official implementation of DigitalCousins~\cite{dai2024acdc}, GPT-4o is used, which requires \$1.25 per scene. Our method costs \$0.47 per scene using Gemini 3 Flash.

\subsection{Ablation Study}
\label{supp:sec:expablate}

\noindent\textbf{Scene Canonicalization.}
\cref{supp:fig:exp_ablation_canon,supp:fig:scene_upfix} show the visual comparisons before and after scene canonicalization. 
We observe that canonicalization (Canon. $\sceneinit$) significantly reduces large pose and orientation errors in the original SAM3D predictions. However, the resulting scenes are still physically unstable due to heavy object intersections. \\

\noindent\textbf{SAM3D with Scene Tree.}
Since SAM3D$\dagger$ lacks explicit spatial reasoning, its reconstructed scenes $\sceneraw$ often contain unsupported objects (\eg, wall-mounted or ceiling-hanging objects) that fall during simulation. 
To isolate whether this failure is solely caused by such objects, we further evaluate SAM3D$\dagger$ combined with our inferred scene tree $\scenetree$ during physics simulation. As shown in \cref{supp:tab:sam3dscenetree}, introducing the scene tree leads to a noticeable improvement in stability. However, due to remaining interpenetrations in the reconstructed geometry, the system is still highly unstable. We further apply the same scene tree $\scenetree$ to the canonicalized scene $\sceneinit$, which yields additional but limited improvements.

Overall, these results indicate that the instability of SAM3D is not only caused by wall- or ceiling-attached objects, and that simple scene canonicalization cannot fully resolve geometric and physical inconsistencies. In contrast, our physics-constrained optimization significantly improves stability across all metrics. \\

\begin{table}[t]
\centering
\small
\caption{\textbf{Effect of scene tree on SAM3D$\dagger$ and canonicalized scenes.} We evaluate SAM3D$\dagger$ $\sceneraw$ and canonicalized scenes $\sceneinit$ under physics simulation with and without our scene tree $\scenetree$.}
\label{supp:tab:sam3dscenetree}
\resizebox{\linewidth}{!}{
\setlength{\tabcolsep}{0.6mm}
\begin{tabular}{lccccc}
\toprule
{Method} & Coll. Rate$\downarrow$ & Stable Rate$\uparrow$ & Pos. Drift$\downarrow$ & Lin. Vel.$\downarrow$ & Ang. Vel.$\downarrow$ \\
\midrule
SAM3D $\dagger$ ($\sceneraw$) & 45.1 & 12.0 & 1.111 & 1.939 & 10.246 \\
SAM3D $\dagger$ ($\sceneraw$) + $\scenetree$ & 45.1 & 36.6 & 0.714 & 1.195 & 6.916 \\
Canon. ($\sceneinit$) & 26.3 & 45.8 & 0.767 & 1.227 & 9.724 \\
Canon. ($\sceneinit$) + $\scenetree$ & 26.3 & 70.6 & 0.334 & 0.488  & 2.902 \\
\midrule
\textbf{Ours ($\scene$)} & \textbf{2.7} & \textbf{94.9} & \textbf{0.072} & \textbf{0.138} & \textbf{0.880} \\
\bottomrule
\end{tabular}}
\end{table}

\begin{table}[t]
\centering
\caption{\textbf{Ablation on energy terms.} We ablate different terms in \cref{eq:energy_decomp}.}
\vspace{-0.8em}
\label{tab:ablation_reward}
\resizebox{\linewidth}{!}{%
\setlength{\tabcolsep}{0.6mm}
\begin{tabular}{l cccccc cc}
\toprule
Ablation
 & Fail. Rate$\downarrow$ 
 & Coll. Rate$\downarrow$ 
  & Stable Rate$\uparrow$
  & Pos. Drift$\downarrow$
  & Lin. Vel.$\downarrow$
  & Ang. Vel.$\downarrow$
  & CD$\downarrow$
  & F-Score$\uparrow$ \\
\midrule
\textit{w/o} $\lossstab$   & 5.0      &   4.4   & 90.8  &  0.136  &  0.260  &  1.423 & {0.018} & 0.820   \\
\textit{w/o} $\lossvel$    & \textbf{0.0} &   5.5   & 92.8  &  0.109  &  0.168  &  0.798 & 0.017 & 0.823         \\
\textit{w/o} $\losscoll$   & \textbf{0.0} &   9.6    & 92.2  &  0.103  &  0.179  &  1.320 & \textbf{0.016} & \textbf{0.828}     \\
\textit{w/o} $\losslay$    & \textbf{0.0} &   \textbf{2.2}    & 91.8  &  {0.075}  &  0.144  &  \textbf{0.778} &  0.024 & 0.726  \\
\textbf{Ours}              & \textbf{0.0} &   2.7   & \textbf{94.9}     &   \textbf{0.072}  &  \textbf{0.138}  &  {0.880}  &   0.017 & 0.824        \\
\bottomrule
\end{tabular}}
\end{table}

\noindent\textbf{Energy terms.}
We ablate different energy terms in our energy function \cref{eq:energy_decomp}. As shown in \cref{tab:ablation_reward}, removing the stability term $\lossstab$ leads to failure in approximately 5\% of cases where stable solutions cannot be obtained. Interestingly, removing the penetration term $\losscoll$ yields the best geometric metrics but reduces physical stability. This is consistent with our analysis in \cref{sec:exp_sota}: since the initial reconstructions often contain interpenetration, removing the collision penalty keeps objects closer to their original positions, improving geometric similarity at the cost of physical plausibility. Without the layout constraint $\losslay$, the optimizer tends to separate heavily intersecting objects to improve stability, causing larger deviations from the original scene layout. Overall, removing any energy term degrades performance, while our full model achieves the best overall results. Please refer to Supp. \cref{supp:sec:expablate} for more ablative experiments.

\begin{figure*}[t]
  \centering
  \includegraphics[width=.96\linewidth]{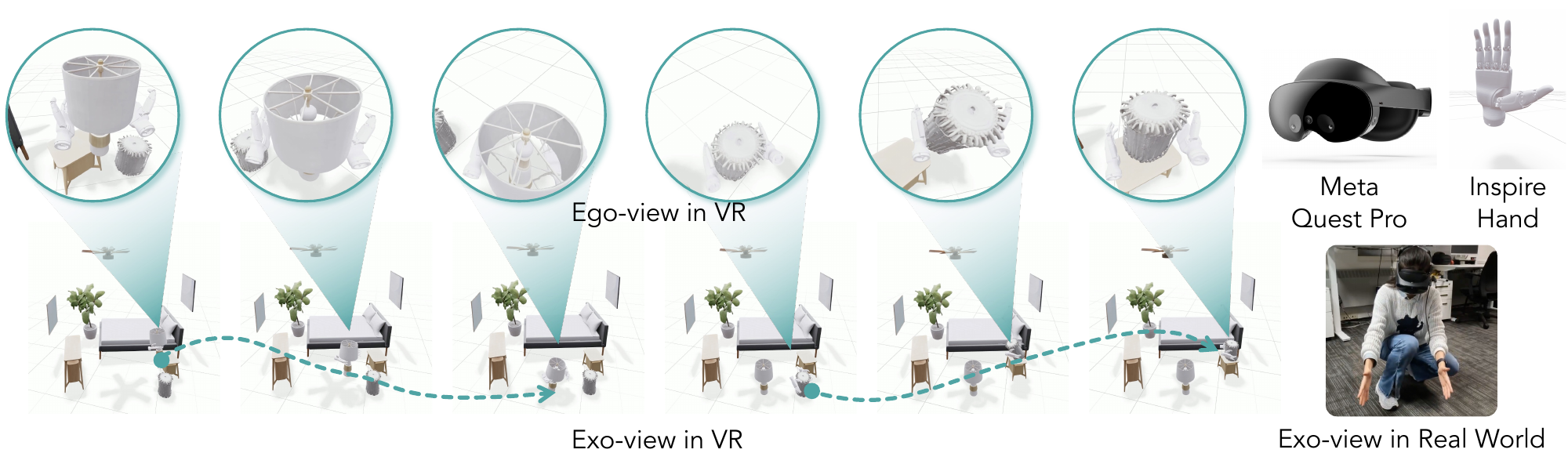}
  \vspace{-0.8em}
  \caption{\textbf{VR interaction setup and demo.} We build our VR-based interaction system using a Meta Quest Pro headset. User hand motions captured via real-time hand tracking are mapped to a simulated Inspire hand in Isaac Gym, enabling physically grounded interaction with reconstructed 3D scenes.}
  \label{supp:fig:supp_vr}
\end{figure*}

\noindent\textbf{CEM elite fraction.}
\cref{supp:tab:ablation_elite_frac} ablates the elite fraction $\rho$, which controls the proportion of top-scoring samples used to update the CEM distribution at each iteration. We observe that very small values of $\rho$ can lead to insufficiently stable updates due to limited elite diversity, while overly large values reduce the selectivity of the optimization process. Overall, $\rho=0.025$ provides a better balance between exploration and exploitation, resulting in more stable and physically consistent solutions. We therefore adopt $\rho{=}0.025$ as the default setting.

\begin{table}[t]
\centering
\caption{\textbf{Ablation on CEM elite fraction.} We study the effect of different elite fraction $\rho$ during CEM optimization.}
\label{supp:tab:ablation_elite_frac}
\resizebox{\linewidth}{!}{%
\setlength{\tabcolsep}{2.4mm}
\begin{tabular}{l cccc}
\toprule
Elite Frac $\rho$ & {Stable Rate} $\uparrow$ & {Pos. Drift} $\downarrow$ & {Lin. Vel.} $\downarrow$ & {Ang. Vel.} $\downarrow$ \\
\midrule
0.01              & 92.0          & 0.081          & 0.204          & 1.047 \\
\textbf{0.025 (Ours)} & \textbf{94.9} & {0.072} & \textbf{0.138} & \textbf{0.880} \\
0.05              & 92.9          & {0.107} & 0.199          & {0.911} \\
0.1               & 91.2          & \textbf{0.060}          & 0.192          & 1.087 \\
0.2               & 90.2          & 0.109          & 0.231          & 1.223 \\
0.5               & 83.0          & 0.124          & 0.276          & 1.535 \\
\bottomrule
\end{tabular}}
\end{table}

\subsection{Real-world VR Interaction}
\label{supp:sec:vr}

\begin{figure}[t]
  \centering
  \includegraphics[width=\linewidth]{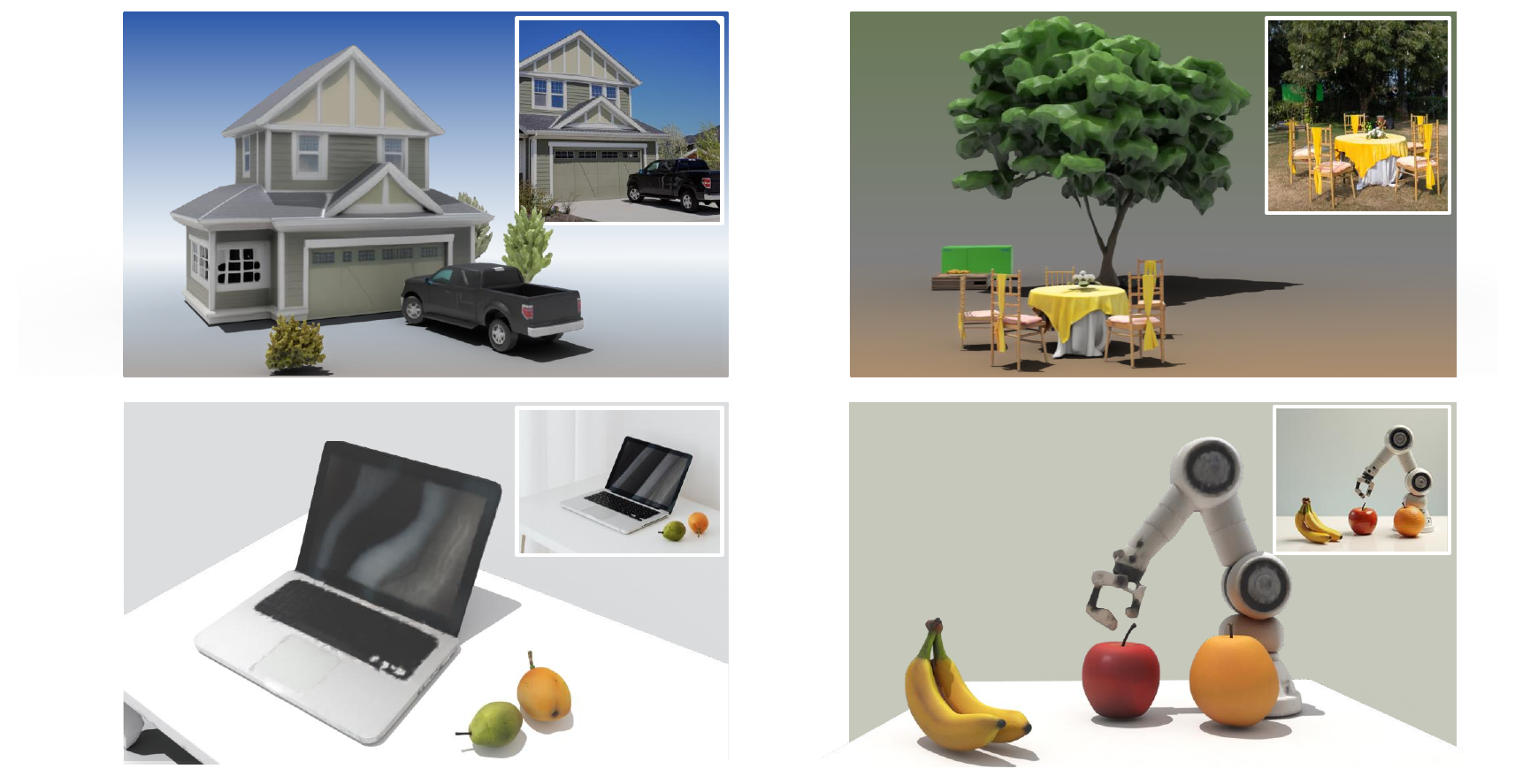}
    \caption{\textbf{Reconstructions on more scenes.} The top row shows tabletop scenes and non-rigid objects, such as laptops and robot arms, while the bottom row shows outdoor scenes.}
  \label{supp:fig:exp_tabletop_arti}
\end{figure}

\begin{figure*}[t]
  \centering
  \includegraphics[width=.93\linewidth]{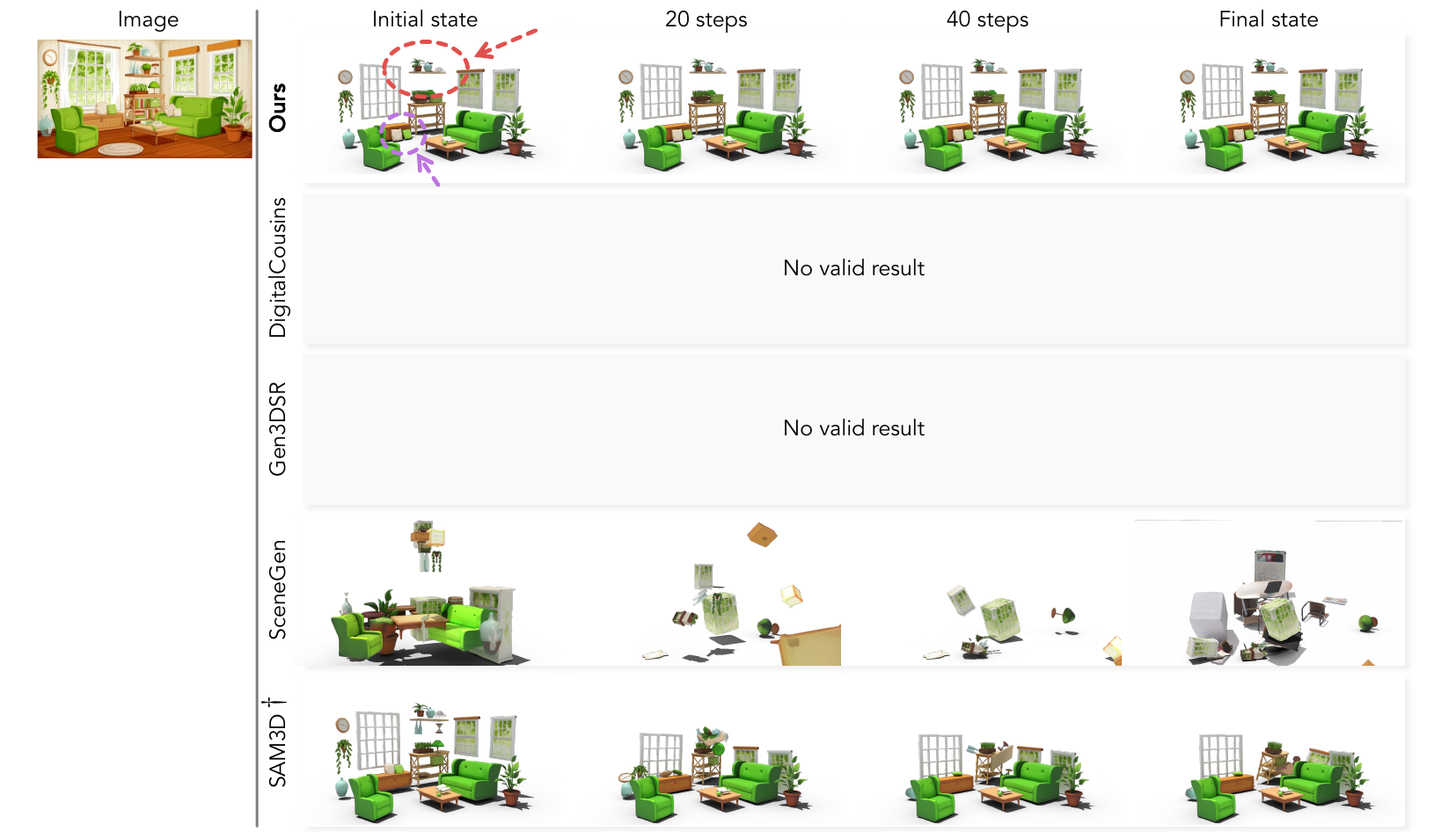}
  \vspace{-0.8em}
  \caption{\textbf{A typical limitation case analysis.} We show a qualitative comparison of simulation results with prior methods. Due to open-vocabulary detection failure, \ie missing a wall-mounted shelf (purple dashed circle), our reconstruction may deviate from the input image. In addition, walls are not explicitly modeled during optimization, causing some wall-supported objects such as the two pillows on a bench (red dashed circle) to be relocated to the ground for physical stability. Despite these issues, our method still significantly outperforms prior approaches.}
  \label{supp:fig:failure}
\end{figure*}

\noindent\textbf{System overview.}
We build a VR interaction system that connects a reconstructed 3D scene with a physics-based simulator, \ie Isaac Gym \cite{makoviychuk2021isaac}, enabling users to manipulate objects in real time using a VR headset, as shown in \cref{supp:fig:supp_vr}. We formulate VR interaction as a physics-based control problem, where user hand motions are continuously mapped to two simulated dexterous Inspire Hands that interact with the scene under rigid-body dynamics. The system runs in a real-time closed loop (30 FPS), where hand tracking data from a Meta Quest Pro headset is streamed to the physics simulator and the updated scene state is rendered back to the user, enabling physically consistent interaction. Please refer to our supplementary video and project website for more VR demonstrations. \\

\noindent\textbf{Hand tracking and control.}
Following Open-TeleVision \cite{cheng2024opentelevision}, we adopt a two-level control scheme for dexterous hand manipulation. Wrist motion is controlled via proportional--derivative (PD) force--torque actuation for physically grounded interaction. Finger joints are obtained using an inverse kinematics-based retargeting method~\cite{qin2023anyteleop}, which maps tracked fingertip positions to feasible joint configurations.

At each simulation step, we compute the position and orientation error between the target wrist pose from VR tracking and the current Isaac wrist state, and apply PD control to generate force and torque commands on the base link. We use a position gain of $200\,\mathrm{N/m}$ with a force clamp of $35\,\mathrm{N}$, and an orientation gain of $10\,\mathrm{Nm/rad}$ with a torque clamp of $5\,\mathrm{Nm}$. In addition, a soft-grip spring is activated when at least two finger links are within $0.12\,\mathrm{m}$ of an object, with stiffness $60\,\mathrm{N/m}$ and damping $8\,\mathrm{N{\cdot}s/m}$. \\

\noindent\textbf{Physics simulator parameters.}
We use a simulation timestep of $1/30\,\mathrm{s}$ with 4 substeps under GPU PhysX. Frictions are empirically set. Object friction is set to $2.0$ for both static and dynamic bodies; combined with hand-link friction $2.0$ under PhysX multiply mode, this yields an effective contact friction of approximately $4.0$, sufficient for stable grasping. Rolling and torsional friction are set to $0.1$. Wall-attached objects use contact and rest offsets of $0.01\,\mathrm{m}$ and $0.005\,\mathrm{m}$, while dynamic objects use $0.005\,\mathrm{m}$ and $0.001\,\mathrm{m}$. Linear and angular damping are set to $8$ and $3$, respectively. Varying these within reasonable ranges doesn’t significantly affect results. Collision geometry is approximated using V-HACD decomposition with a voxel resolution of $10^6$ and up to 32 convex hulls per object. The hull counts are kept low to balance simulation accuracy and runtime. Increasing hull counts shows minimal impact, but longer runtime. \\

\noindent\textbf{Object mass assignment.}
To enable physically realistic grasping, we assign each object a mass based on its semantic category inferred from the scene tree $\scenetree$. The scene representation provides an object list with category names and attributes, which we use to group objects into three classes: \textit{heavy} (sofas, desks, beds, cabinets, \etc.) with mass clamped to $[60, 200]\,\mathrm{kg}$; \textit{medium} (chairs, ottomans, lamps, TVs, etc.) with mass in $[3, 7]\,\mathrm{kg}$; and \textit{light} (cups, books, vases, remote controls, etc.) with mass in $[2.5, 4]\,\mathrm{kg}$. The mass value is set by the \ac{vlm} according to common-sense knowledge of real-world objects. Inertia tensors are scaled proportionally when the natural V-HACD mass is clamped to the category range. 

\subsection{Limitations and Future Directions}
\label{supp:sec:expdiscuss}
Despite significantly improving the physical stability of scene reconstruction, real-world environments remain highly complex, with diverse geometries and open-vocabulary challenges, which expose the limitations of current \ac{vlm}s in visual reasoning. As shown in \cref{supp:fig:failure}, missing detections, such as a shelf mounted on the wall (highlighted by purple dashed circle), may cause the inferred layout to deviate from the original image. Nevertheless, our physics-constrained optimization is still able to find plausible and stable placements for these densely arranged objects. We also note that non-planar grounds are not currently handled; fitting planar primitives \cite{wang2026contactguided} may be a promising future direction.

While our current framework mainly focuses on rigid objects in indoor scenes, it can be flexibly extended to outdoor scenes by adapting the scene analysis prompts to detect structures such as buildings, as demonstrated in \cref{supp:fig:exp_tabletop_arti}, opening up potential applications in autonomous driving. We also observe reasonable robustness on tabletop scenes and scenes containing non-rigid and articulated objects such as laptop or robot arms in \cref{supp:fig:exp_tabletop_arti}. However, articulated objects are currently treated as rigid, and their articulation is not explicitly modeled. Extending the framework to better handle non-rigid and articulated objects is an important direction toward more general and realistic scene reconstruction, with potential applications in downstream robotics and embodied AI tasks.

Another limitation is that we do not explicitly model walls as supporting structures during the physics-constrained optimization. As a result, objects that rely on wall support may not preserve their original layouts after optimization. For instance, in \cref{supp:fig:failure} (highlighted by red dashed circle), two pillows placed on a bench near the window are relocated to the ground, since the wall is not explicitly represented as a support constraint in the optimization process. This optimized solution yields a lower energy state under physical constraints, with the object orientations remaining consistent with the original scene, and is therefore considered more stable. Nevertheless, our method still significantly outperforms existing approaches in terms of physical plausibility and stability. There also remain many promising future directions to explore, including incorporating explicit scene structure priors, such as walls and support surfaces, into the optimization process to better preserve global layout consistency while maintaining physical plausibility.

\end{document}